\documentclass[letterpaper, 10 pt, conference]{ieeeconf}  % Comment this line out if you need a4paper

\usepackage{times}
\usepackage{epsfig}
\usepackage{graphicx}
\usepackage{amsmath}
\usepackage{amssymb}
\usepackage{booktabs}
\usepackage{multirow}
\usepackage[utf8]{inputenc}
\usepackage[T1]{fontenc}
\usepackage[ruled,lined]{algorithm2e}
\usepackage[pagebackref,breaklinks,colorlinks]{hyperref}

\usepackage{neuralnetwork}

\usepackage[capitalize]{cleveref}
\crefname{section}{Sec.}{Secs.}
\Crefname{section}{Section}{Sections}
\Crefname{table}{Table}{Tables}
\crefname{table}{Tab.}{Tabs.}

\IEEEoverridecommandlockouts                              % This command is only needed if 
\title{\LARGE \bf
SOccDPT: Semi-Supervised 3D Semantic Occupancy from Dense Prediction Transformers trained under memory constraints
}

\author{Aditya Nalgunda Ganesh$^{1}$%
\thanks{$^{1}$Aditya Nalgunda Ganesh is with Department of Computer Science and Engineering,
        PES University, India
        {\tt\small adityang5@gmail.com}}%
}

\begin{document}

\maketitle
\thispagestyle{empty}
\pagestyle{empty}

%%%%%%%%%%%%%%%%%%%%%%%%%%%%%%%%%%%%%%%%%%%%%%%%%%%%%%%%%%%%%%%%%%%%%%%%%%%%%%%%
\begin{abstract}

We present SOccDPT, a memory-efficient approach for 3D semantic occupancy prediction from monocular image input using dense prediction transformers. To address the limitations of existing methods trained on structured traffic datasets, we train our model on unstructured datasets including the Indian Driving Dataset and  Bengaluru Driving Dataset. Our semi-supervised training pipeline allows SOccDPT to learn from datasets with limited labels by reducing the requirement for manual labelling by substituting it with pseudo-ground truth labels to produce our Bengaluru Semantic Occupancy Dataset. This broader training enhances our model's ability to handle unstructured traffic scenarios effectively. To overcome memory limitations during training, we introduce patch-wise training where we select a subset of parameters to train each epoch, reducing memory usage during auto-grad graph construction. In the context of unstructured traffic and memory-constrained training and inference, SOccDPT outperforms existing disparity estimation approaches as shown by the RMSE score of 9.1473, achieves a semantic segmentation IoU score of 46.02\% and operates at a competitive frequency of 69.47 Hz. We make our code and semantic occupancy dataset public.

\end{abstract}

%%%%%%%%%%%%%%%%%%%%%%%%%%%%%%%%%%%%%%%%%%%%%%%%%%%%%%%%%%%%%%%%%%%%%%%%%%%%%%%%
\section{INTRODUCTION}

\begin{figure*}[t]
\begin{center}
    \vspace{0.5cm}
   \includegraphics[width=0.3\linewidth]{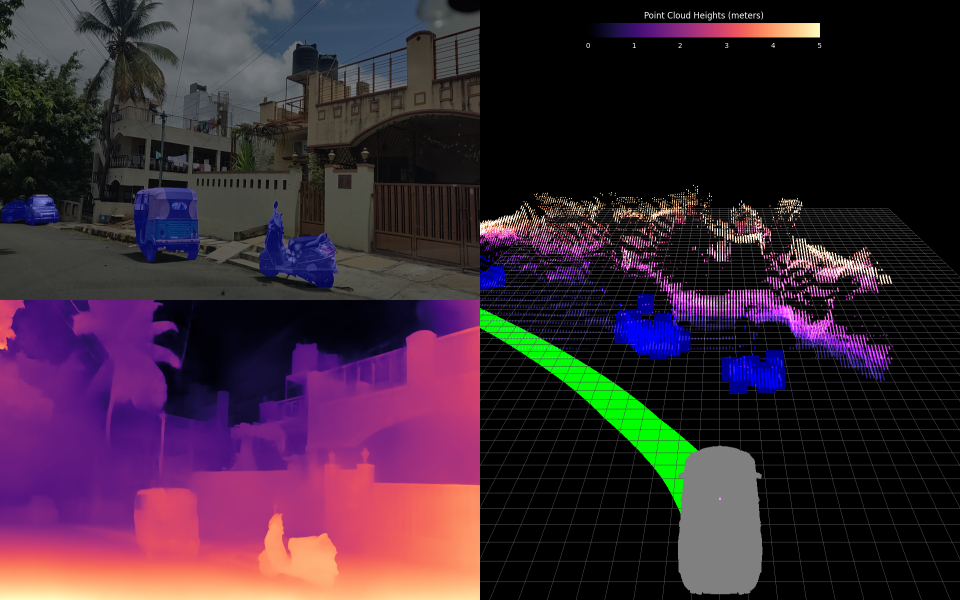}
   \includegraphics[width=0.3\linewidth]{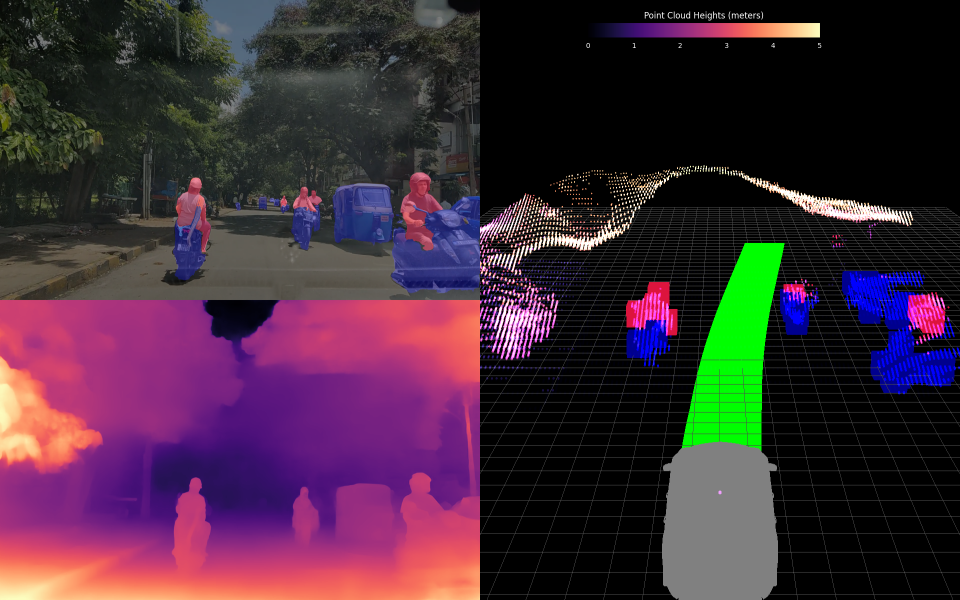}
   \includegraphics[width=0.3\linewidth]{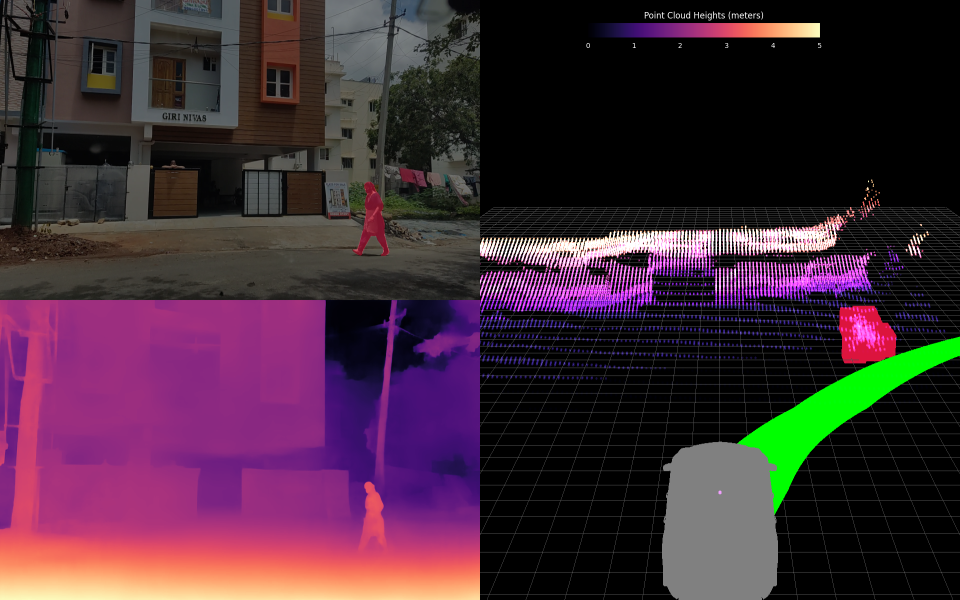} \\

   % \hspace{0.1cm} \\

   % \includegraphics[width=0.33\linewidth]{images/BSOD_frames/temporal/1_visual_20230820_095822.png}
   % \includegraphics[width=0.33\linewidth]{images/BSOD_frames/temporal/4_visual_20230820_095825.png}
   % \includegraphics[width=0.33\linewidth]{images/BSOD_frames/temporal/8_visual_20230820_095830.png} \\

   \includegraphics[width=0.3\linewidth]{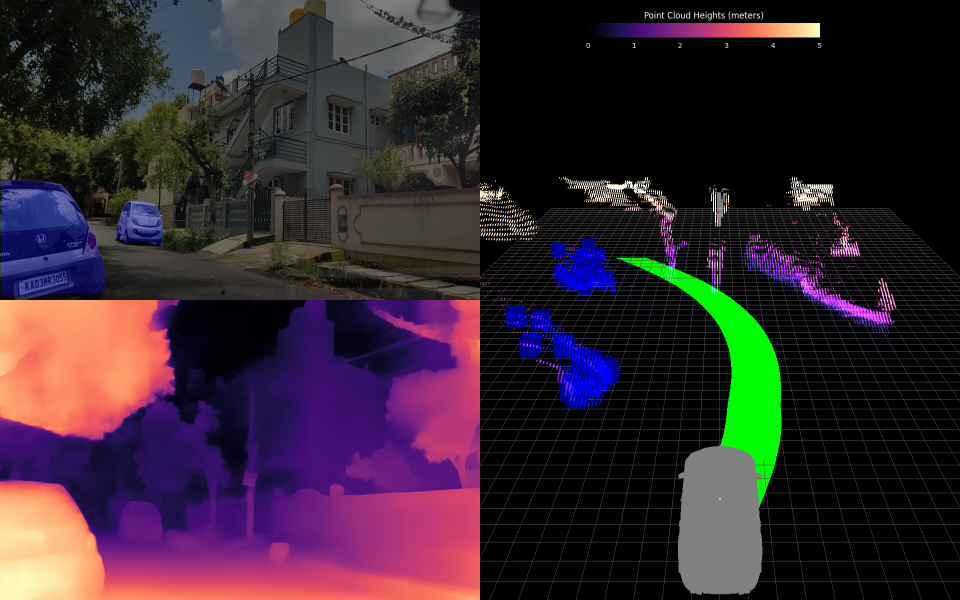}
   \includegraphics[width=0.3\linewidth]{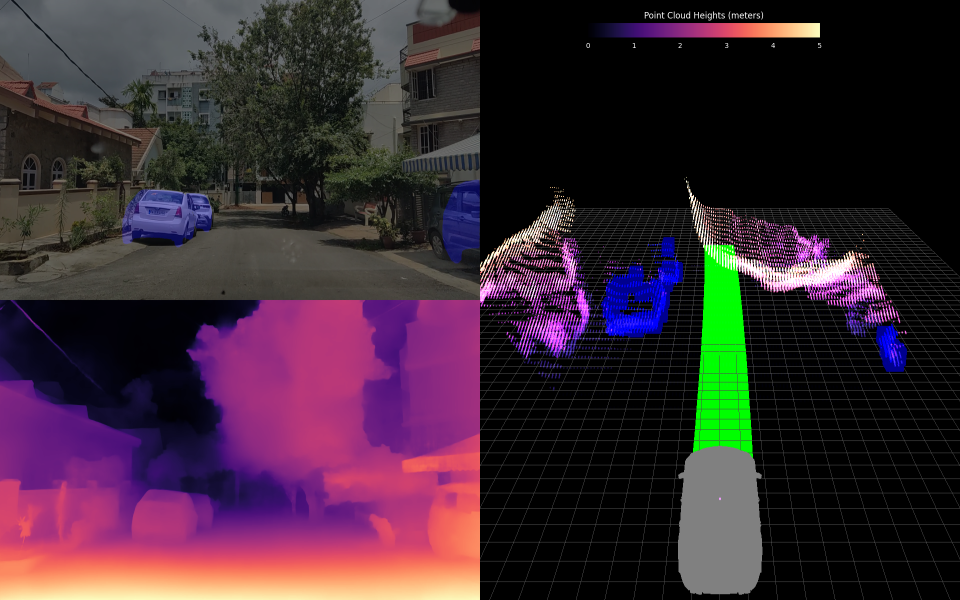}
   \includegraphics[width=0.3\linewidth]{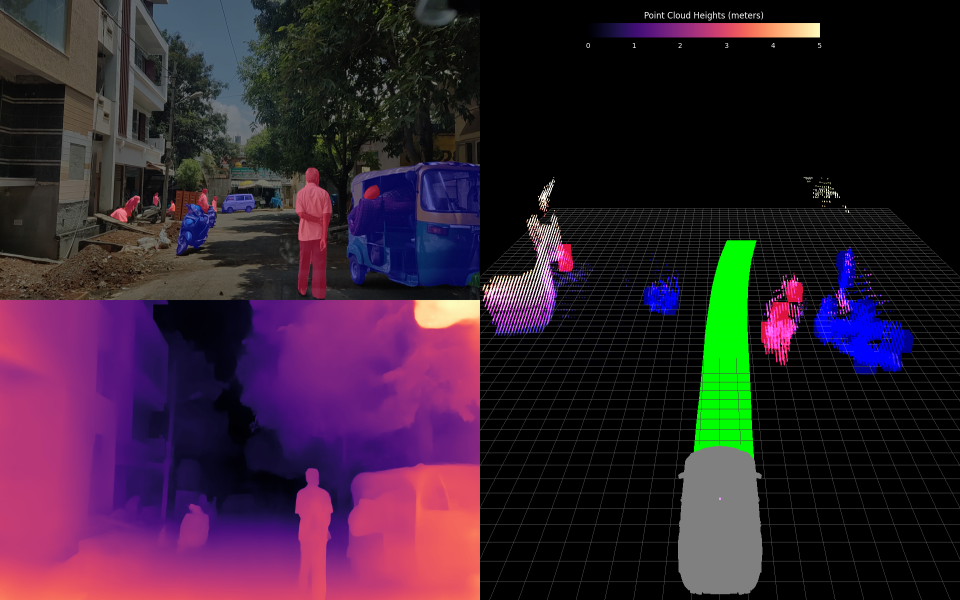} \\

\end{center}
    % \hspace{0.5cm} \\
   \caption{Above are a few frames from our Bengaluru Semantic Occupancy Dataset which is an extension of the Bengaluru Driving Dataset \cite{OCTraN_analgund2023octran}. Each panel consists of the RGB image with 2D semantic labels on the top left, the disparity map on the bottom left and the 3D semantic occupancy on the right. The vehicle and pedestrian classes are colored in blue and red respectively. Objects without classes have been plotted as a height map for the sake of visualization. The vehicle and its future trajectory have been plotted out in grey and green respectively to aid the reader to have a better scene understanding.}
\label{fig:bsod_frames}
\end{figure*}

Autonomous navigation requires 3D semantic understanding of the environment at a high frequency with a limited compute budget. The field of autonomous driving has shown significant interest in vision-based 3D scene perception due to its exceptional efficiency and abundant semantic information. When it comes to choosing an architecture, works such as ~\cite{bao2022beit, Graham_2021_ICCV_levit, jaegle2021perceiver, li2022next_vit, liu2021swinv2, liu2021Swin} inspired from ViT~\cite{ViT} have the domain agnostic learning capabilities of the transformer. The transformer's versatility comes at the cost of having no good inductive priors for any domain, requiring large volumes of training data and a large volume of GPU memory to train. To apply such models on a new domain, we must be efficient in making use of transfer learning and pseudo-labelling to solve the ground truth data scale problem.

In the context of 3D semantic occupancy from monocular vision, ground truth data would refer to semantically labelled 3D point clouds with corresponding RGB images acquired from a calibrated camera sensor as shown in ~\cref{fig:bsod_frames} of our Bengaluru Semantic Occupancy Dataset. While there exist datasets \cite{fong2021_nue_panoptic, KITTI, KITTI_360_DBLP:journals/corr/abs-2109-13410} which have labelled 3D semantic occupancy data in the context of structured traffic, the unstructured traffic scenarios remain largely underrepresented. It may not be feasible to gather large volumes of training data considering the fact that LiDAR sensors are expensive and labelling 3D semantic classes can be tedious. Hence, we make use of a set of teacher models and boosting techniques inspired from~\cite{cheng2021pointly_pointrend, OCTraN_analgund2023octran, bmd_msc_boosting, Miangoleh2021Boosting} to produce labels for depth and semantic on driving video footage which we use to supervise the training of our model. We train our system on unstructured driving datasets such as the Indian Driving Dataset~\cite{IDD_DBLP:journals/corr/abs-1811-10200} and Bengaluru Driving Dataset~\cite{OCTraN_analgund2023octran} to ensure that our system generalizes well. Training such models requires large volume of GPU memory. We overcome this hurdle with our PatchWise training approach which keeps the GPU memory in check and this allowed us to explore higher batch sizes without altering the back-propagation algorithm.

With the goal of designing a model, which is efficient during both training and inference, we propose SOccDPT and our PatchWise training system. To ensure SOccDPT performs well in unstructured traffic scenarios, we introduce semi-supervision with our pseudo-labelling process for depth boosting and semantic auto-labelling. We use a common backbone for image feature extraction and dual heads to extract disparity and semantic information of the scene. Camera intrinsics are used along with disparity to project the semantic information into 3D space. 
% Memory constrained training
% https://arxiv.org/pdf/2111.11124.pdf

\section{RELATED WORK}

\begin{figure*}[t]
\begin{center}
   \includegraphics[width=1.0\linewidth]{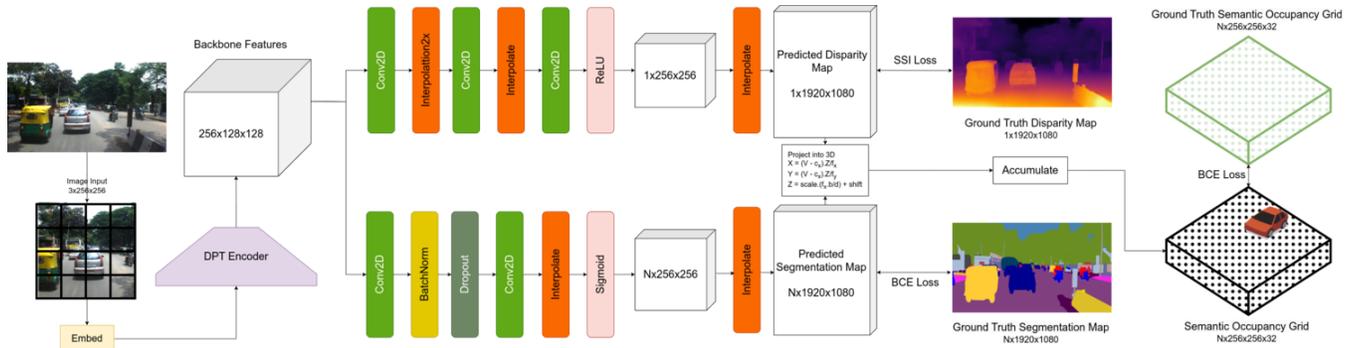}
\end{center}
   \caption{SOccDPT uses the ViT family for backbone feature extraction which allows us to carefully balance accuracy and compute requirements. SOccDPT takes an RGB image input of shape $3 \times 256 \times 256$ produces image features of shape 256x128x128. We then pass the extracted features to a disparity head and a segmentation head. We apply the Scale and Shift Invariant loss~\cite{Ranftl2022_midas_ssi_loss} and the Binary Cross Entropy loss for the disparity and segmentation outputs respectively. With the known camera intrinsic, we project the semantics into 3D space with the help of the disparity map and accumulate the semantics into a 3D occupancy grid of size $256 \times 256 \times 32$, thus producing a 3D semantic map from one backbone}
\label{fig:architecture}
\end{figure*}

\textbf{Multi-Task Learning. } The domain of visual perception has extensively studied the concept of Multi-Task Learning including tasks such as semantic segmentation \cite{Liu2018ERN, joint_semantic_mtl} and jointly learning depth and semantics ~\cite{9706962_multi_task_ssl}. A comprehensive discussion of deep multi-task learning can be found in ~\cite{mtl_survey}. Although there have been some advancements in semi-supervised multi-task learning, as evidenced by previous studies~\cite{Cheng2016SemiSupervisedMD, 6963417_ssl_mtl_scene}, these approaches do not specifically tackle the difficult problem of training models across diverse datasets in the absence of ground truth information. Our focus is to address this gap by developing a solution for semi-supervised multi-task learning in the domain of 3D semantic occupancy within unstructured traffic environments.
% ~\cite{9706962_multi_task_ssl}

\textbf{Semi-Supervised Learning and Self-Supervised Learning. }
In the context of disparity estimation, semi-supervised learning has become very important due to the challenges involved in obtaining accurate depth information in diverse real-world environments. Several self-supervised algorithms for perceiving depth have been suggested \cite{monodepth, monodepth2, megadepth, gcndepth, manydepth}. These algorithms offer the advantage of utilizing only a single camera, making them suitable for easy deployment in real-world scenarios. However, they still face numerous unresolved issues. One such problem is the generation of disparity maps that lack local and temporal consistency. Watson et al. \cite{manydepth} addressed the temporal inconsistency by incorporating multiple consecutive frames as input. Another line of research in semi-supervised learning looks into using the existing model to generate confident annotations on unlabelled data. Examples for such approaches include pseudo-labelling~\cite{hung2018adversarial_pseudo_label} and entropy minimization~\cite{NIPS2004_96f2b50b_entropy_min}. Since the degree of disparity is inversely related to depth, as demonstrated in \cref{eq:project_3d}, slight variations in disparity for distant objects lead to significant variations in depth. Consequently, the resulting point clouds exhibit non-uniform resolution, with closer objects represented by more points compared to those farther away. There are broadly two approaches tot he disparity estimation problem: monocular and stereoscopic.

\textbf{Monocular and Stereo Disparity Estimation. } Diverse neural network architectures, including variational auto-encoders, convolutional neural networks, generative adversarial networks and recurrent neural networks, have demonstrated their efficacy in tackling the task of depth estimation. Within this framework, two methods are commonly employed: monocular, where depth is estimated from a single input image, and stereoscopic depth estimation, where depth is estimated from a pair of images provided as input to the system. Monocular approaches such as ~\cite{Ranftl2022_midas_ssi_loss, monodepth, monodepth2, manydepth} take advantage of depth cues such as occlusion boundaries, parallel lines and so on to understand the 3D scene. Techniques based on binocular learning \cite{cheng2020hierarchical, UASNet, UPFNet, xu2023iterative_IGEVStereo}, generate depth maps by leveraging the epipolar constraints associated with feasible disparity values. Although these approaches have enhanced accuracy, they come with a trade-off in terms of computational time and/or hardware demands. Running such systems in real-time becomes impractical on embedded devices that lack sufficient power, especially without specialized hardware like FPGAs.

\textbf{Bird's Eye View (BEV) Architectures.} Since disparity is inversely proportional to depth, errors in depth estimation grow quadratically with disparity errors. To address these errors, some approaches look to operate in the Bird's Eye View space which is directly proportional to depth.  Obtaining a top-down view of a scene offers a comprehensive understanding of the surrounding environment, effectively capturing both static and dynamic elements. BEV architectures, exemplified by \cite{li2022bevformer, reiher2020sim2real, roddick2020predicting}, generate this top-down map, which can be utilized for path planning purposes. This top down map is essentially a segmentation map which would highlight the road, non-drivable space, parking areas, vehicles, pedestrians and so on. The concept of predicting BEV from multiple camera perspectives has demonstrated performance comparable to LiDAR-centric methods \cite{liu2022bevfusion, zhu2020cylindrical}. However, a limitation of this approach is the absence of 3D information about the scene, such as unclassified objects, potholes, and overhanging obstacles.

\textbf{3D Occupancy Networks.} Difficult to classify 3D obstacles become harder and harder to catch as the research community and industry chases the long tail of nines. Recent approaches look to build generic 3D object detection free of ontology by the way of 3D Occupancy Networks. Achieving an effective representation of a 3D scene is a fundamental objective in perceiving 3D environments. One direct approach involves discretizing the 3D space into voxels within an occupancy grid \cite{zhou2017voxelnet, zhu2020cylindrical}. The voxel-based representation is advantageous for capturing intricate 3D structures, making it suitable for tasks like LiDAR segmentation \cite{cheng2021af2s3net, liong2020amvnet, tang2020searching, ye2022lidarmultinet, ye2021drinet, zhu2020cylindrical} and 3D scene completion \cite{cao2022monoscene, chen20203d, li2020anisotropic, roldão2020lmscnet, yan2020sparse}. A recent method, TPVFormer \cite{huang2023triperspective}, addresses memory optimization by representing the 3D space as projections on three orthogonal planes. Despite the significant progress made by these approaches, they do not specifically tackle the issue of existing dataset biases towards structured traffic.

% \begin{figure}
%     \centering
%     % \includegraphics{
%     \begin{neuralnetwork}[height=2]
%         \newcommand{\x}[2]{$x_#2$}
%         \newcommand{\y}[2]{$\hat{y}_#2$}
%         \newcommand{\hfirst}[2]{\small $h^{(1)}_#2$}
%         \newcommand{\hsecond}[2]{\small $h^{(2)}_#2$}
%         \inputlayer[count=1, bias=true, title=Input\\layer, text=\x]
%         \hiddenlayer[count=1, bias=true, title=Hidden\\layer 1, text=\hfirst] \linklayers
%         % \hiddenlayer[count=1, bias=true, title=Hidden\\layer 2, text=\hsecond] \linklayers
%         \outputlayer[count=1, title=Output\\layer, text=\y] \linklayers
%     \end{neuralnetwork}
%     % }

%     \hspace{0.5cm} \\
    
%     \caption{A perceptron which takes one input, produces one output and has a single hidden layer}
%     \label{fig:mlp_example}
% \end{figure}

\bgroup

\section{PROPOSED WORK}

\begin{figure}
    \vspace{0.5cm}
    \centering
    \begingroup
    \newcommand*\rot{\rotatebox{90}}
    \setlength{\tabcolsep}{1pt} % Default value: 6pt
    \renewcommand{\arraystretch}{0.5} % Default value: 1

  \begin{tabular}{lll}
\rot{\hspace{4.5mm}{\tiny RGB 
\textcolor{white}{[]} % necessary to align all the columns
}}
\includegraphics[width=.30\columnwidth]{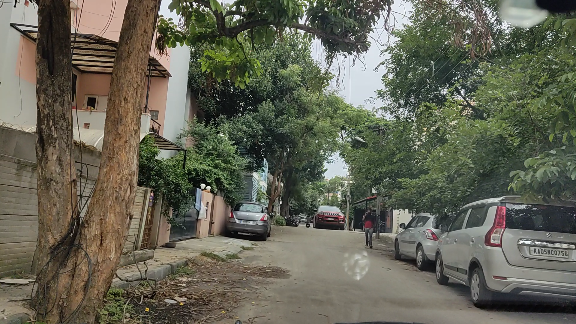} & 
\includegraphics[width=.30\columnwidth]{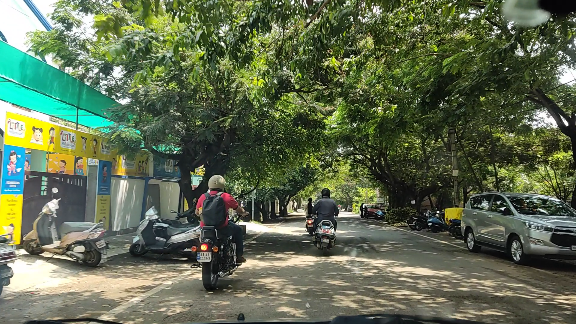} &
\includegraphics[width=.30\columnwidth]{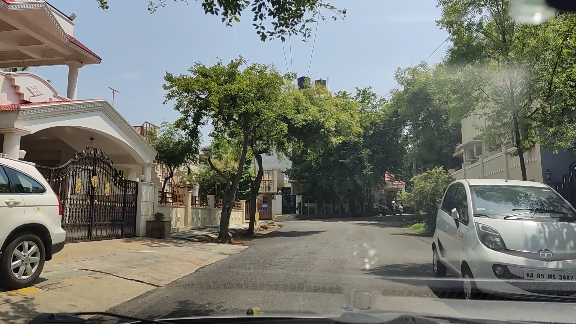}\\

\rot{\hspace{2.0mm}{\tiny Ground Truth
}}
\includegraphics[width=.30\columnwidth]{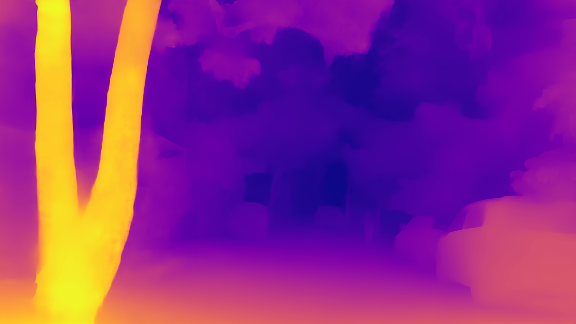} & 
\includegraphics[width=.30\columnwidth]{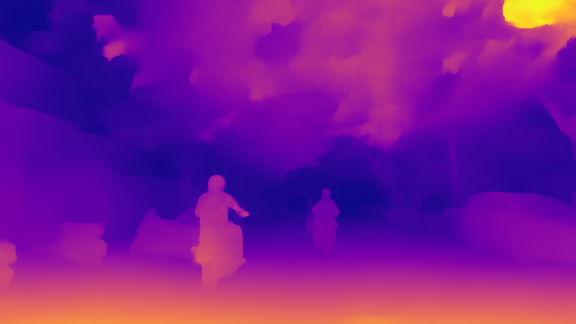} &
\includegraphics[width=.30\columnwidth]{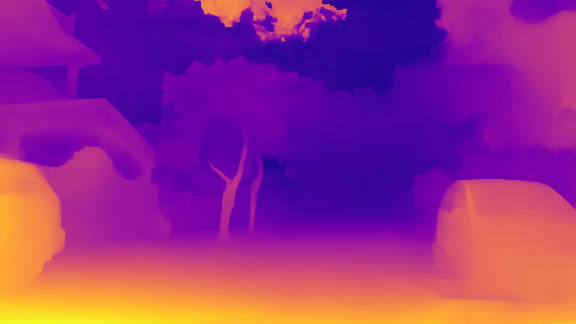}\\

\rot{\hspace{1mm}{\tiny $SOccDPT$
}}
\includegraphics[width=.30\columnwidth]{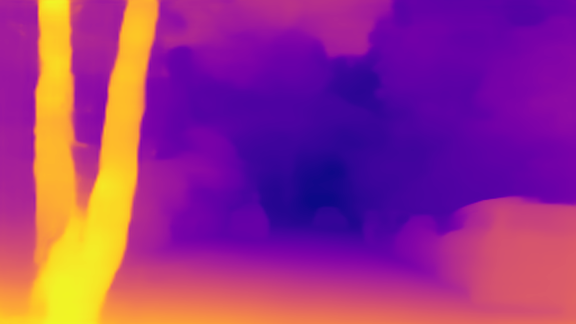} & 
\includegraphics[width=.30\columnwidth]{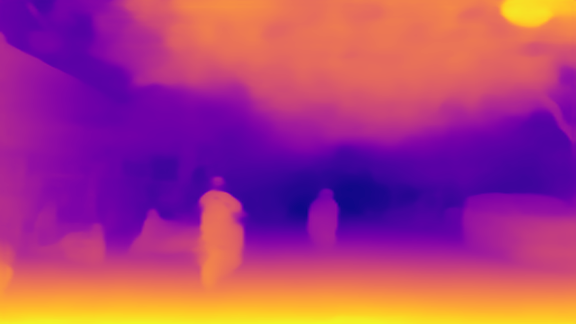} &
\includegraphics[width=.30\columnwidth]{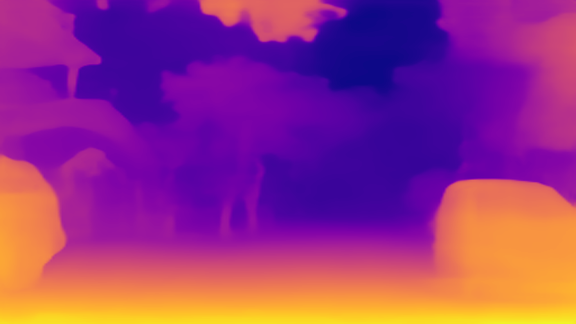}\\

\rot{\hspace{1mm}{\tiny $MiDaS$\cite{Ranftl2020_DPT}
}}
\includegraphics[width=.30\columnwidth]{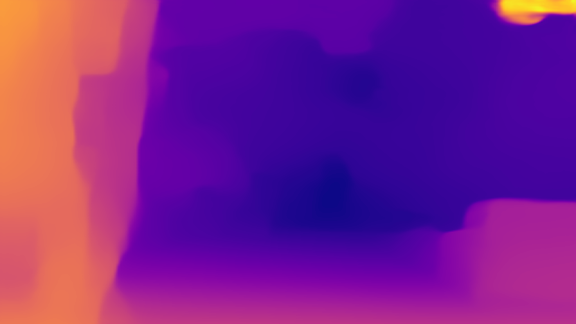} & 
\includegraphics[width=.30\columnwidth]{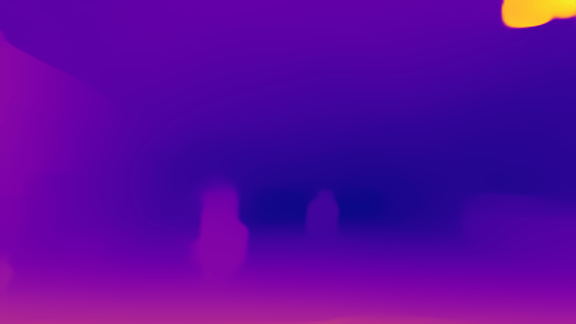} &
\includegraphics[width=.30\columnwidth]{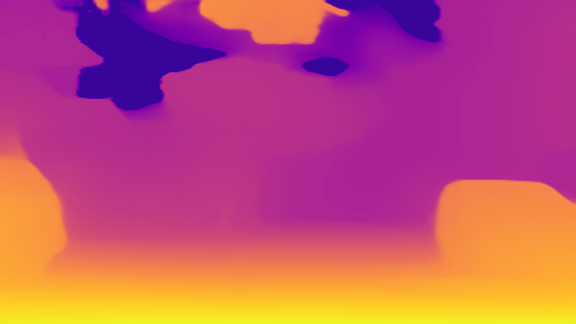}\\

\rot{\hspace{1mm}{\tiny Manydepth \cite{manydepth}
}}
\includegraphics[width=.30\columnwidth]{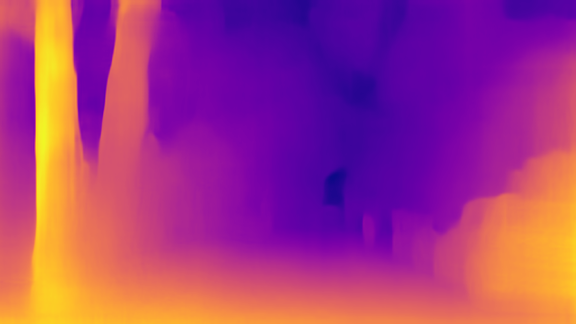} & 
\includegraphics[width=.30\columnwidth]{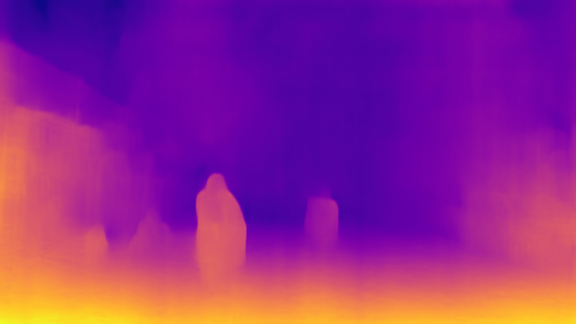} &
\includegraphics[width=.30\columnwidth]{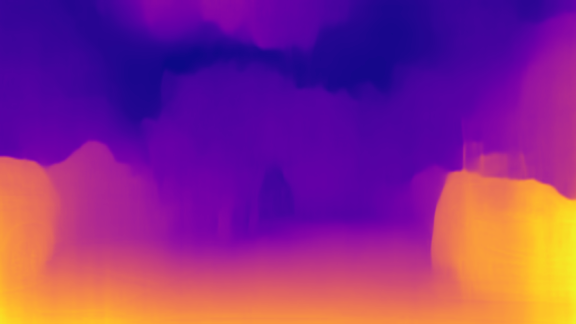}\\

% \rot{\hspace{1mm}{\tiny Monodepth2 \cite{monodepth2}
% }}
% \includegraphics[width=.30\columnwidth]{images/qualitative/0_5.png} & 
% \includegraphics[width=.30\columnwidth]{images/qualitative/1_5.png} &
% \includegraphics[width=.30\columnwidth]{images/qualitative/2_6.png}\\

\rot{\hspace{2mm}{\tiny PackNet \cite{tri-packnet}
}}
\includegraphics[width=.30\columnwidth]{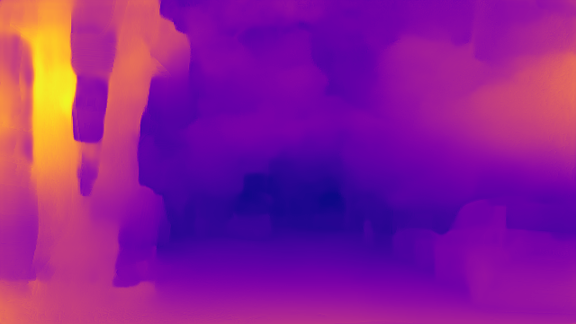} & 
\includegraphics[width=.30\columnwidth]{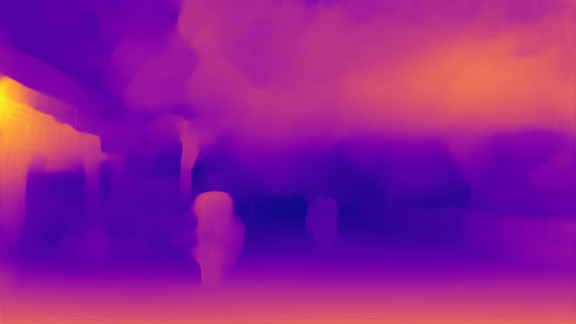} &
\includegraphics[width=.30\columnwidth]{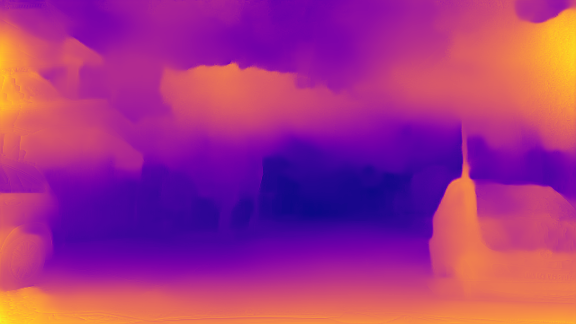}\\

\rot{\hspace{2mm}{\tiny ZeroDepth \cite{tri-zerodepth}
}}
\includegraphics[width=.30\columnwidth]{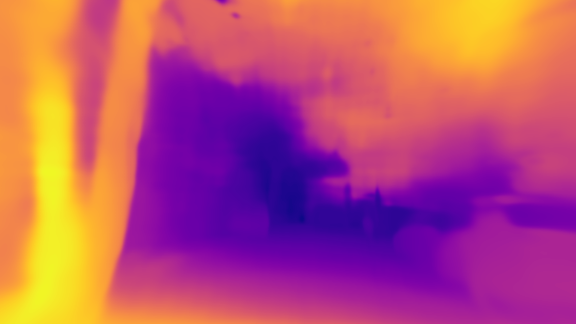} & 
\includegraphics[width=.30\columnwidth]{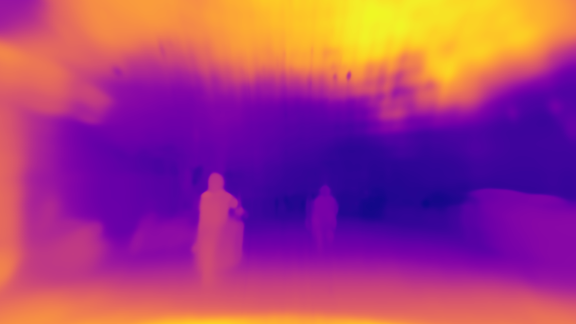} &
\includegraphics[width=.30\columnwidth]{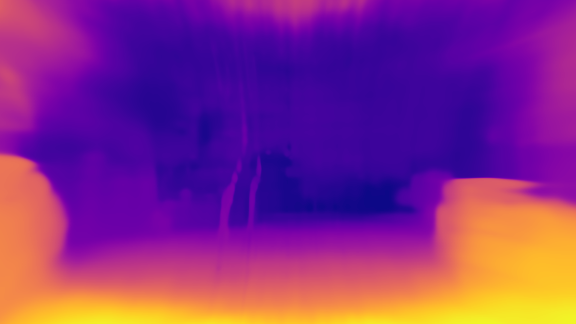}\\

  \end{tabular}
  \endgroup
  \caption{Qualitative results comparing frames in BDD to Midas versions, monodepth, manydepth, ZeroDepth. As we can see, all the existing approaches do not address the diversity that is seen in unstructured traffic}
  \label{fig:qualitative_results} 
\end{figure}

\subsection{SOccDPT Architecture}

As described in ~\cref{fig:architecture}, SOccDPT uses the Dense Prediction Transformer~\cite{Ranftl2020_DPT, liu2021swinv2} backbones to efficiently extract image features. We then use independent heads to produce the disparity and segmentation maps. Instead of penalizing the model for generating the output in an inaccurate scale, we address the issue of arbitrary scale in the disparity map by estimating the scale and shift relative to the ground truth for every frame. This estimation process involves aligning the prediction with the ground truth using a least-squares criterion.

Once the segmentation and disparity maps are computed, we make use of the camera intrinsics to project the semantics into 3D space. Consider a point on the image plane at position $(u,v)$ with disparity $D(u,v)$ and 2D semantics $S_{2D}(u,v)$. This point corresponds to the 3D point $(x,y,z)$ as shown in ~\cref{eq:project_3d} from which we can assert the 3D semantics correspondence to be $S_{3D}(x,y,z) \longleftarrow S_{2D}(u,v)$

\begin{equation}
\begin{split}
(x,y,z) = (\frac{b \cdot (u-o_x)}{D(u,v)}, \frac{b \cdot f_x \cdot (v-o_y)}{f_y \cdot D(u,v)}, \frac{b \cdot f_x}{D(u,v)})
\end{split}
\label{eq:project_3d}
\end{equation}

In order to train our network, we started off by building a baseline model $V1$ which consists of 2 separate backbones, one for disparity and the other for segmentation. This informs us of the performance of the dense prediction transformer on unstructured traffic datasets. We improve upon $V1$ by having a common backbone in $V2$ which lead to optimizations in speed and memory consumption. This came at the cost of the accuracy of both the segmentation and disparity. This is due to the fact that the network would be learning the features and intricacies of both the tasks from scratch simultaneously. To address this, $V3$ makes a minor modification to $V2$ which allows us to load in the disparity estimation backbone from $V1$. This allows $V3$ to have a backbone which is proficient in the disparity estimation task. When starting from this point, the backbone and segmentation head only have to learn the task of image segmentation, without making any major alterations to the existing disparity estimation. This provided an improvement in how much the model was able to learn with the same data.

\subsection{PatchWise Training}

Our PatchWise system, implemented in PyTorch \cite{PyTorch_NEURIPS2019_9015}, offers a solution to GPU memory limitations during neural network training. Instead of updating all weights simultaneously, which can lead to "out of memory" errors, PatchWise updates a subset of the model's weights at a time. This approach enables the training of larger networks and the use of larger batch sizes on systems with limited GPU memory, although it increases training time. The implementation details are described in ~\cref{algo:patchwise}.

\begin{algorithm}
\caption{PatchWise}\label{alg:patchwise}

\SetKwInOut{Input}{Input}
\SetKwInOut{Output}{Output}

\SetKwFunction{PatchWise}{\text{PatchWise}}

\underline{PatchWise} $(\text{net}, \text{train\_percentage}, \text{train\_step})$\;
\Input{PyTorch Module \text{net}, training percentage \text{train\_percentage}, training function \text{train\_step}}
\Output{Trained neural network}

\BlankLine
$\text{N} \gets \text{length}(\text{net.parameters})$\;
$\text{M} \gets \text{round}(\text{N} \times \text{train\_percentage})$\;
$\text{num\_iterations} \gets \lceil \text{N} / \text{M} \rceil$\;
$\text{updated\_weights} \gets \{\}$\;
$\text{saved\_weights} \gets \{\}$\;
\BlankLine
\For{$\text{index}, \text{param} \text{ in }\text{net.parameters}$}{
    $\text{saved\_weights[index]} \gets \text{param}$\;
}

\BlankLine
\For{$\text{net\_patch\_index} \text{ in range}(0, \text{num\_iterations})$}{
    \BlankLine
    % \tcp{Compute the start and end index}
    $\text{start\_index} \gets \text{net\_patch\_index} \times \text{M}$\;
    $\text{end\_index} \gets \min(\text{start\_index} + \text{M}, \text{N})$\;

    % \BlankLine
    % \tcp{Reset all the parameters}
    % \tcp{Unfreeze the next M parameters}
    $\text{train\_indices} \gets \text{range}(\text{start\_index}, \text{end\_index})\;$
    \For{$\text{index}, \text{param} \text{ in }\text{net.parameters}$}{
        $\text{param} \gets \text{saved\_weights[index]}$\;
        
        $\text{param.requires\_grad} \gets \text{bool}($\\
            % \STATE\hspace{\algorithmicindent}
            \hspace{\algorithmicindent}
            $\text{index} \in \text{train\_indices}$ \\
        $)$\;
    }

    \BlankLine
    % \tcp{Train step to get updated weights}
    \text{train\_step}(\text{net})\;

    \BlankLine
    % \tcp{Save the updated parameters}
    $\text{save\_indices} \gets \text{range}(\text{start\_index}, \text{end\_index})$\;
    \For{$\text{index}, \text{param} \text{ in }\text{net.parameters}$}{
        \If{$\text{index} \in \text{save\_indices}$}{
            $\text{updated\_weights[index]} \gets \text{param}$\;
        }
    }
}
\BlankLine
% \tcp{Batch update all parameters}
\For{$\text{index}, \text{param} \text{ in }\text{net.parameters}$}{
    $\text{param} \gets \text{updated\_weights[index]}$\;
}

\BlankLine
\KwRet \text{net}\;

\label{algo:patchwise}
\end{algorithm}

\subsection{Pseudo-Ground Truth Labels for Semi-Supervision}

\begin{figure}
    \centering
    \begingroup
    \newcommand*\rot{\rotatebox{90}}
    \setlength{\tabcolsep}{1pt} % Default value: 6pt
    \renewcommand{\arraystretch}{0.5} % Default value: 1

  \begin{tabular}{lll}
\includegraphics[width=.32\columnwidth]{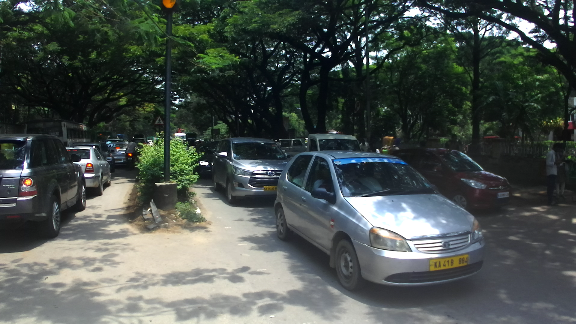} & 
\includegraphics[width=.32\columnwidth]{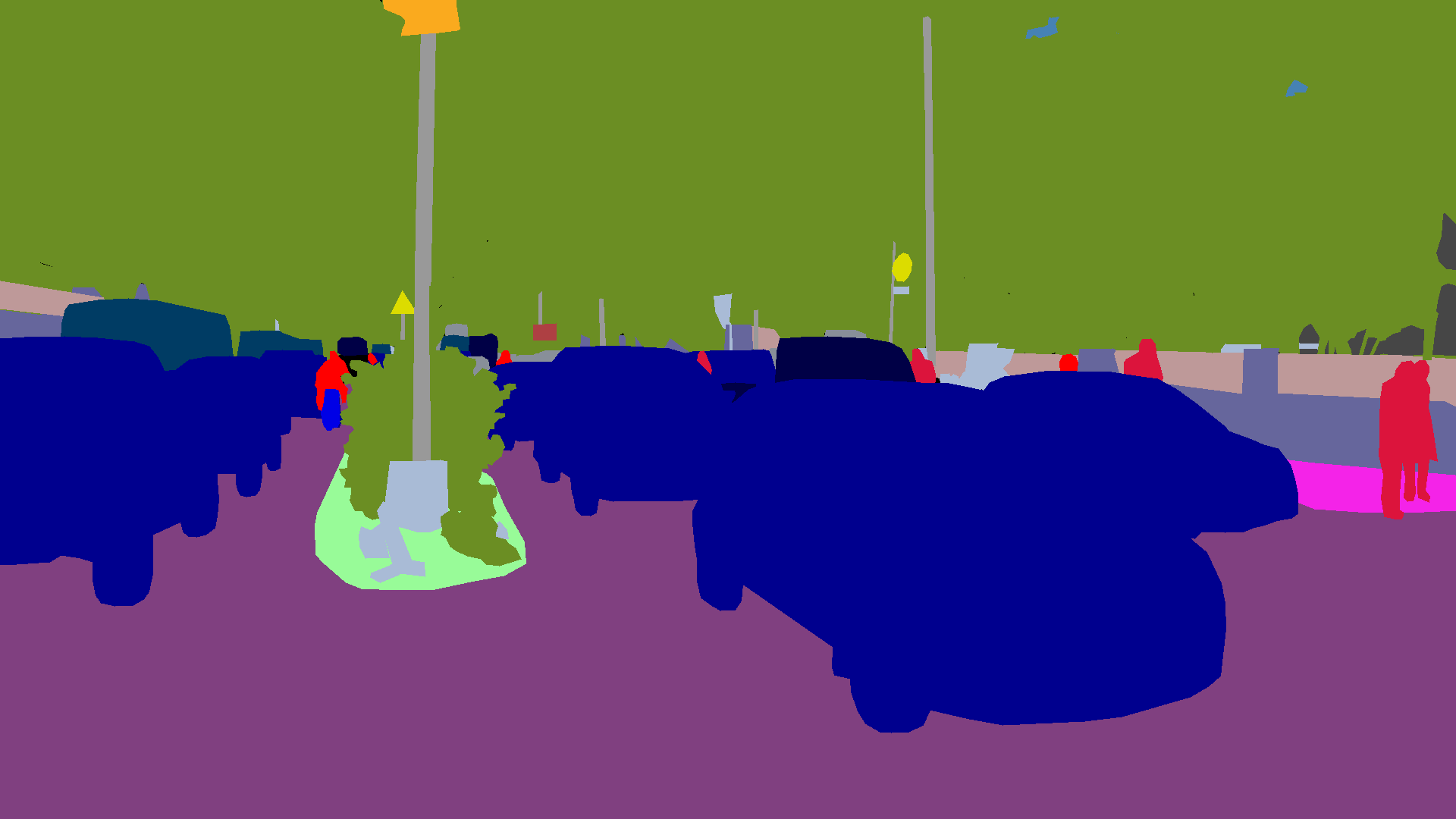} & 
\includegraphics[width=.32\columnwidth]{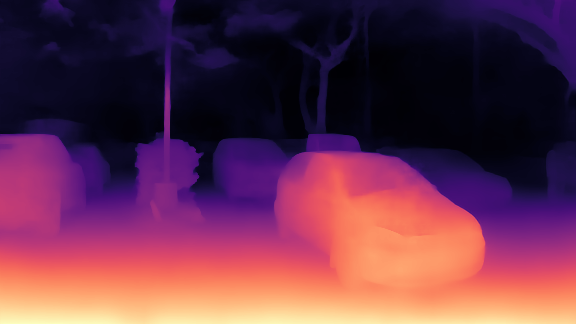}\\

\includegraphics[width=.32\columnwidth]{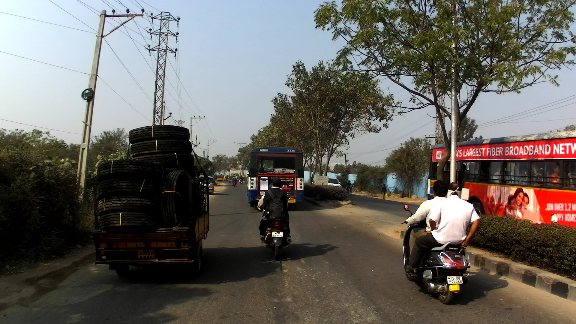} & 
\includegraphics[width=.32\columnwidth]{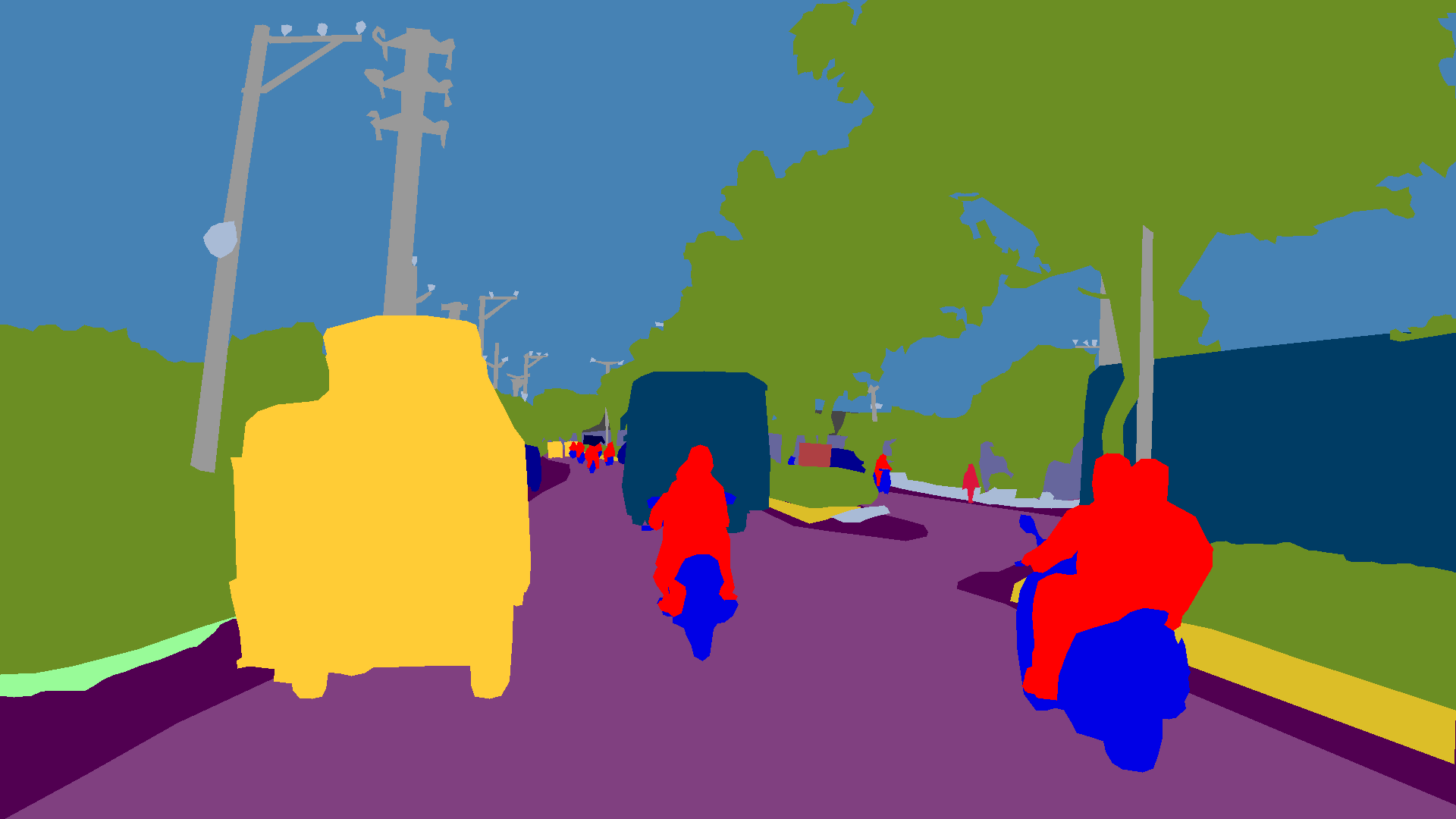} & 
\includegraphics[width=.32\columnwidth]{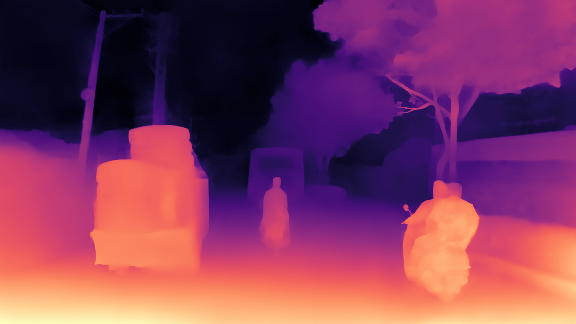}\\

\includegraphics[width=.32\columnwidth]{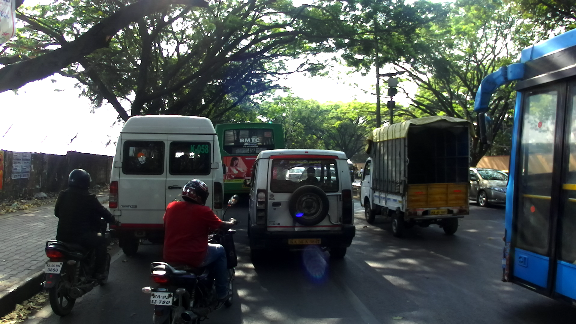} & 
\includegraphics[width=.32\columnwidth]{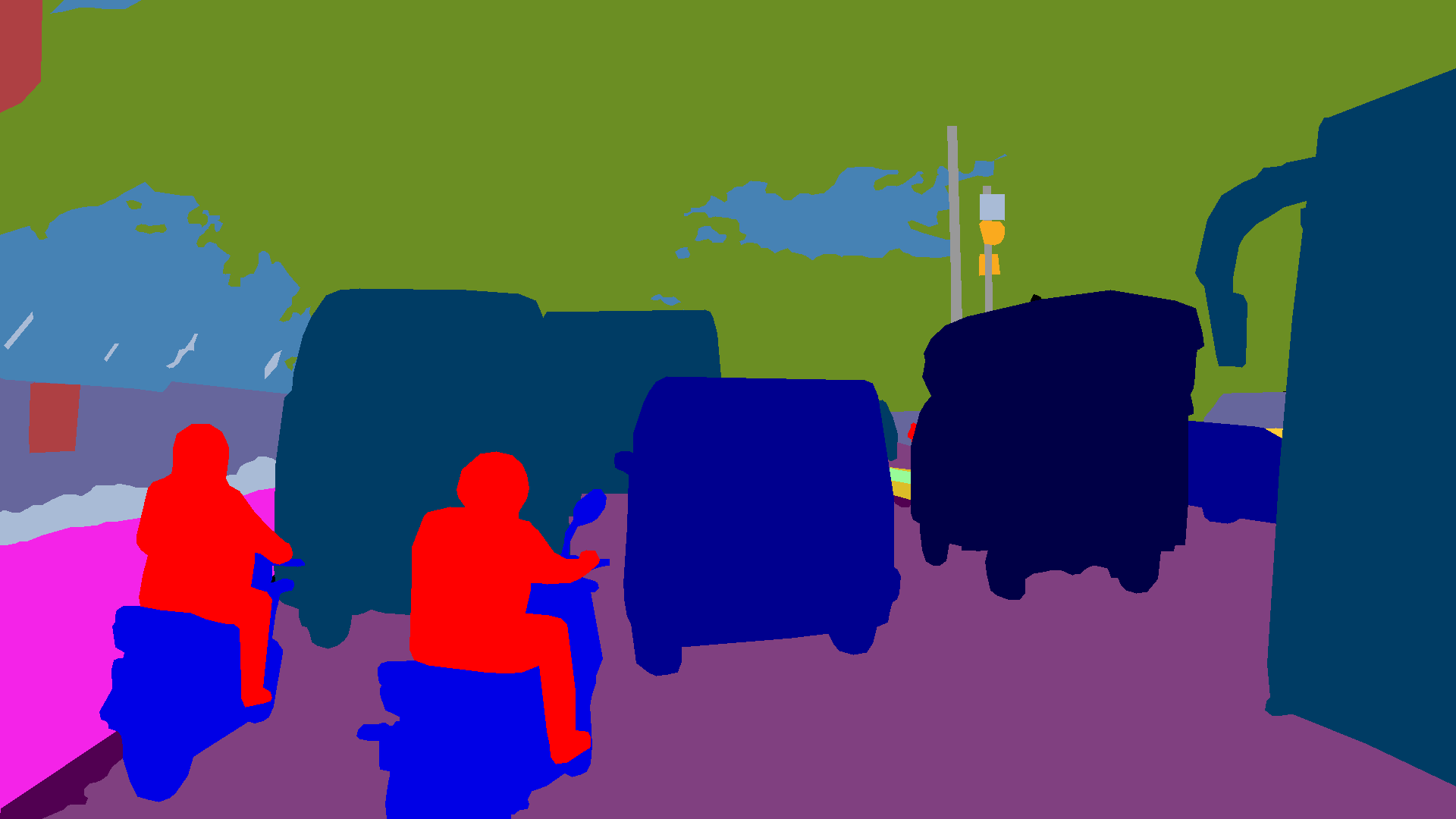} & 
\includegraphics[width=.32\columnwidth]{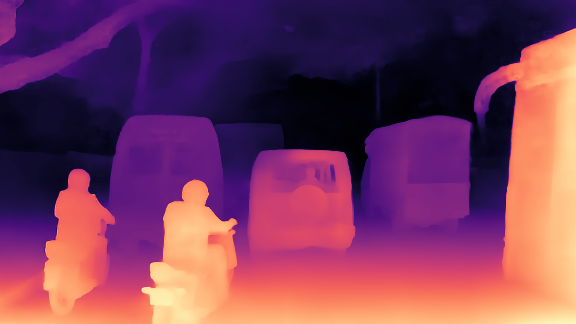}\\

\includegraphics[width=.32\columnwidth]{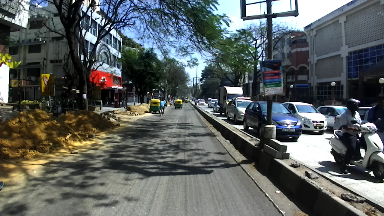} & 
\includegraphics[width=.32\columnwidth]{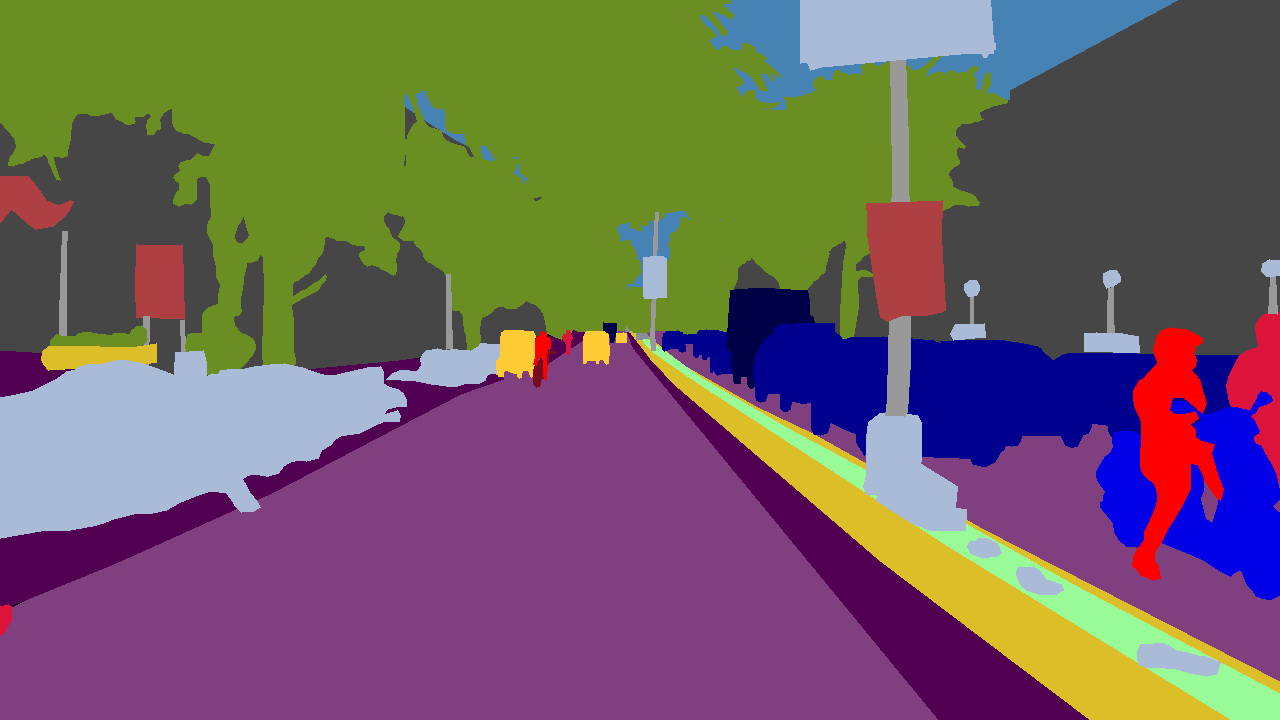} & 
\includegraphics[width=.32\columnwidth]{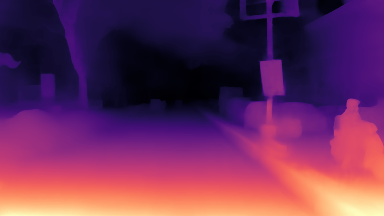}\\

\includegraphics[width=.32\columnwidth]{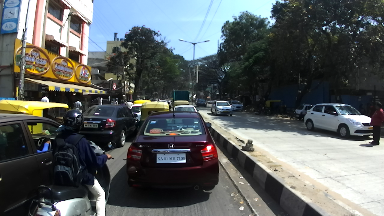} & 
\includegraphics[width=.32\columnwidth]{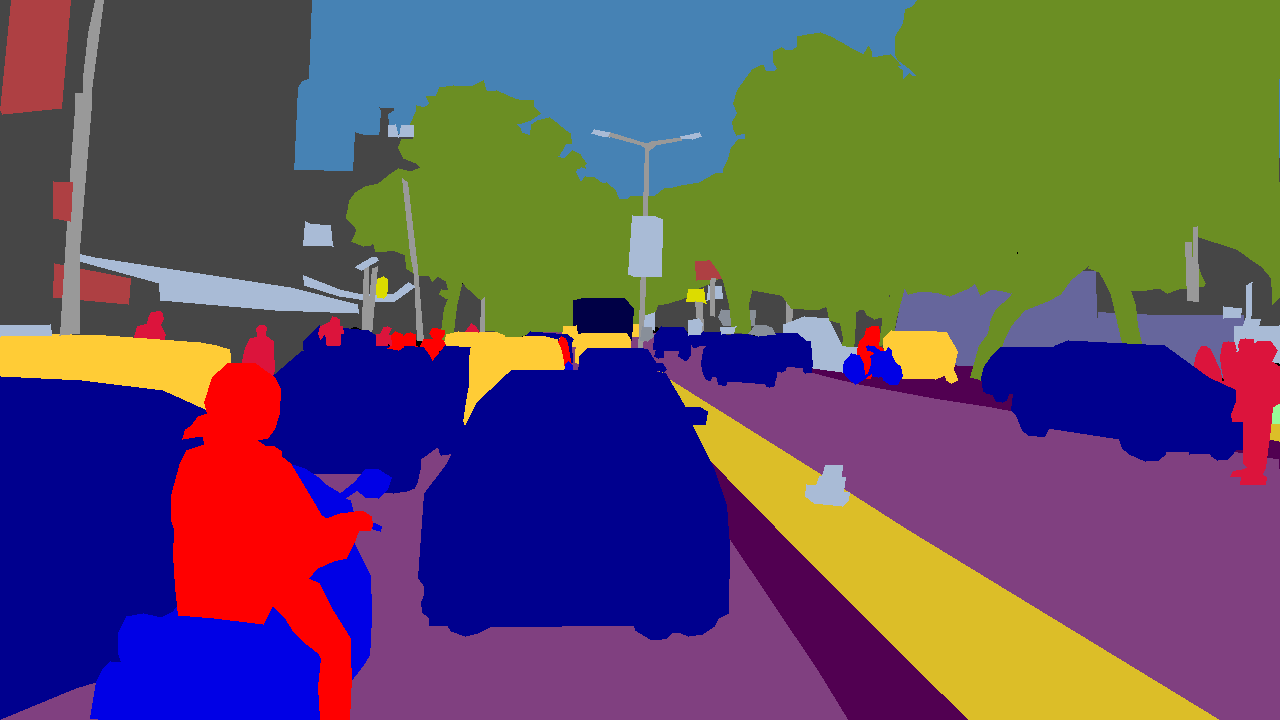} & 
\includegraphics[width=.32\columnwidth]{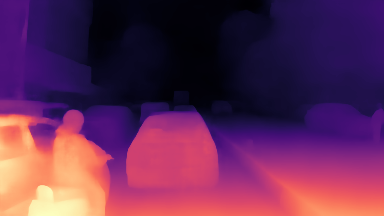}\\

\includegraphics[width=.32\columnwidth]{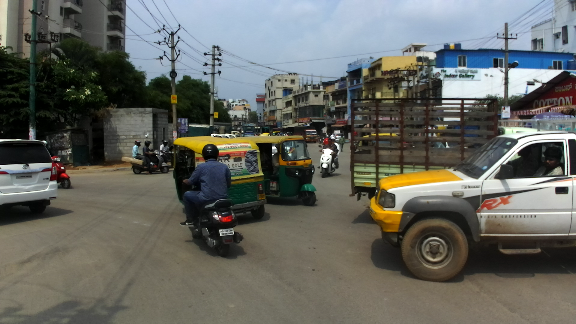} & 
\includegraphics[width=.32\columnwidth]{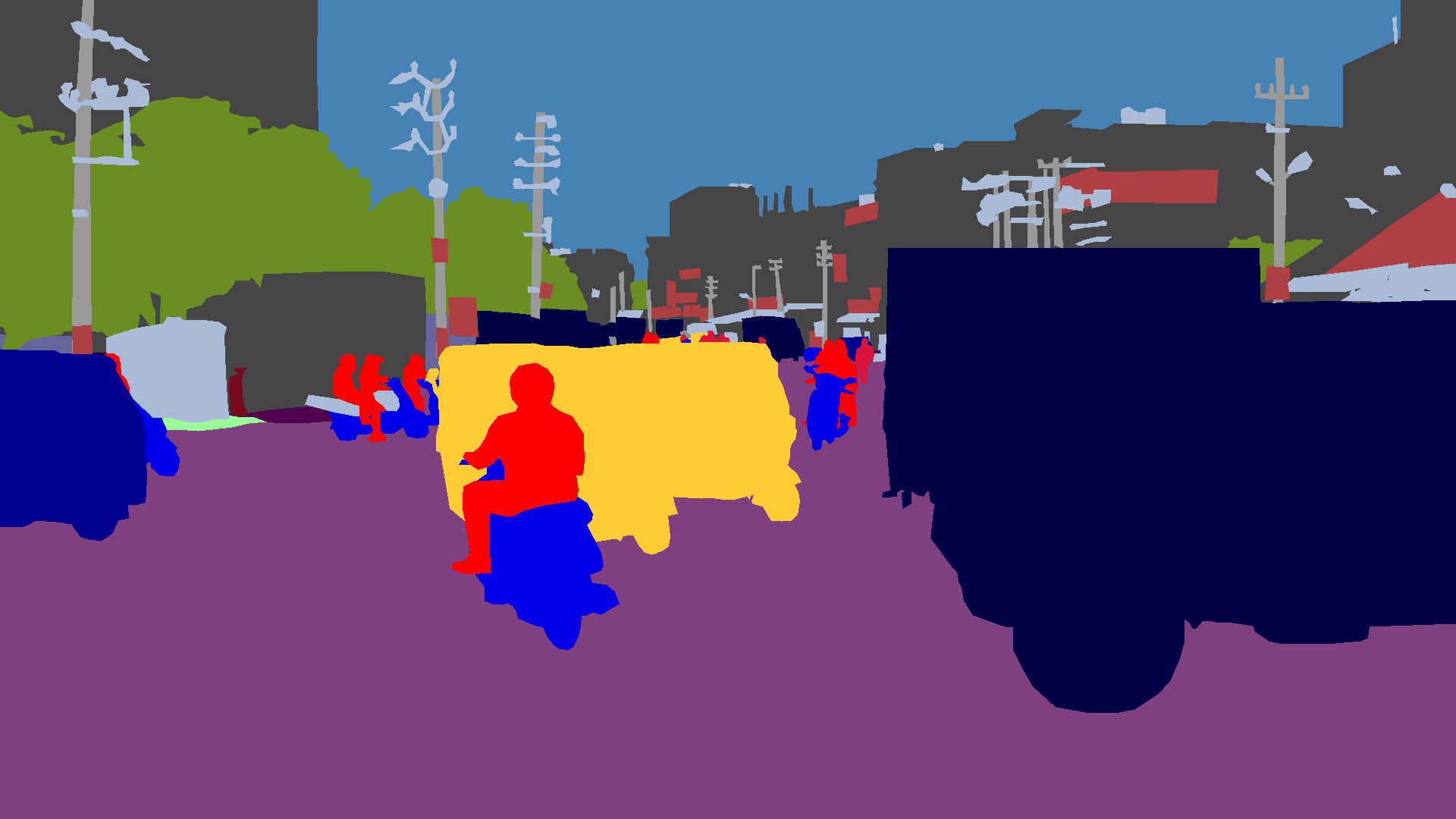} & 
\includegraphics[width=.32\columnwidth]{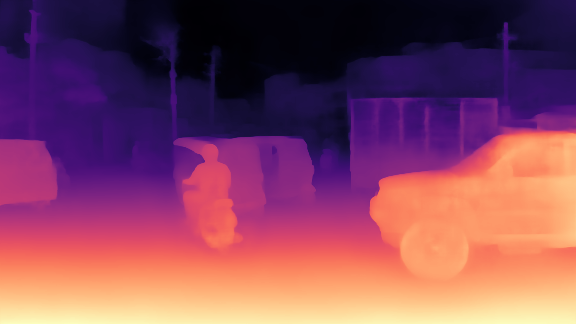}\\

  \end{tabular}
  \endgroup

  \hspace{0.5cm} \\
  
  \caption{We use Depth Boosting to generate depth labels for the Indian Driving Dataset. We have the RGB frames on the left, segmentation map in the middle and our depth labels on the right. We would like to highlight the detail in the automatically generated disparity maps}
  \label{fig:idd_depth_labels} 
\end{figure}

\begin{figure}
    \centering
    \begingroup
    \newcommand*\rot{\rotatebox{90}}
    \setlength{\tabcolsep}{1pt} % Default value: 6pt
    \renewcommand{\arraystretch}{0.5} % Default value: 1

  \begin{tabular}{lll}
\includegraphics[width=.32\columnwidth]{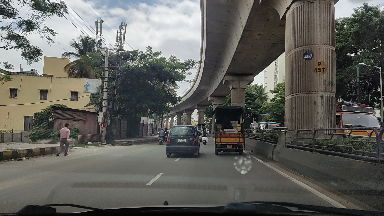} & 
\includegraphics[width=.32\columnwidth]{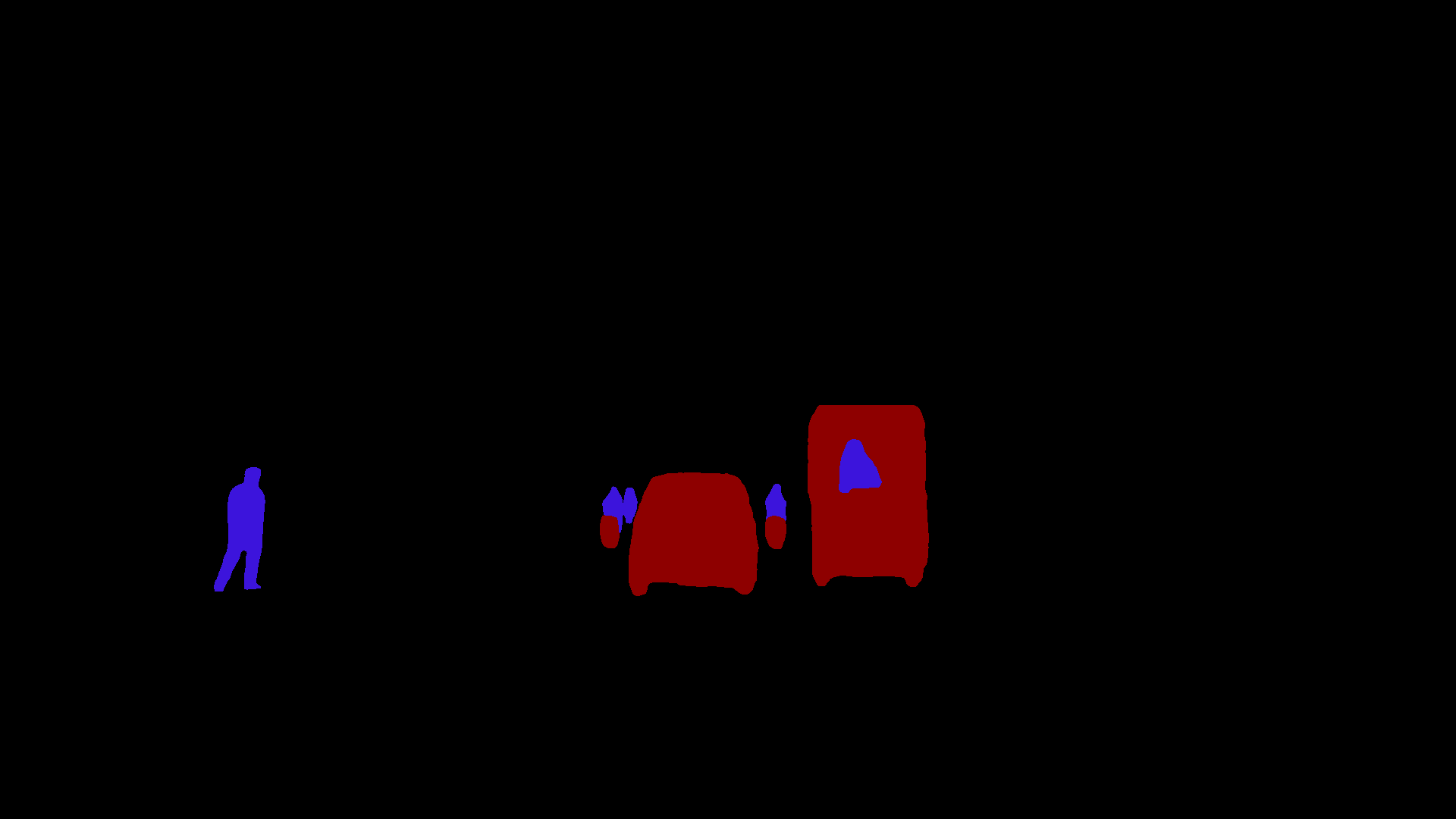} & 
\includegraphics[width=.32\columnwidth]{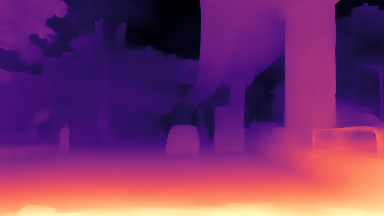}\\

\includegraphics[width=.32\columnwidth]{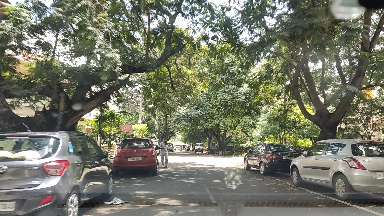} & 
\includegraphics[width=.32\columnwidth]{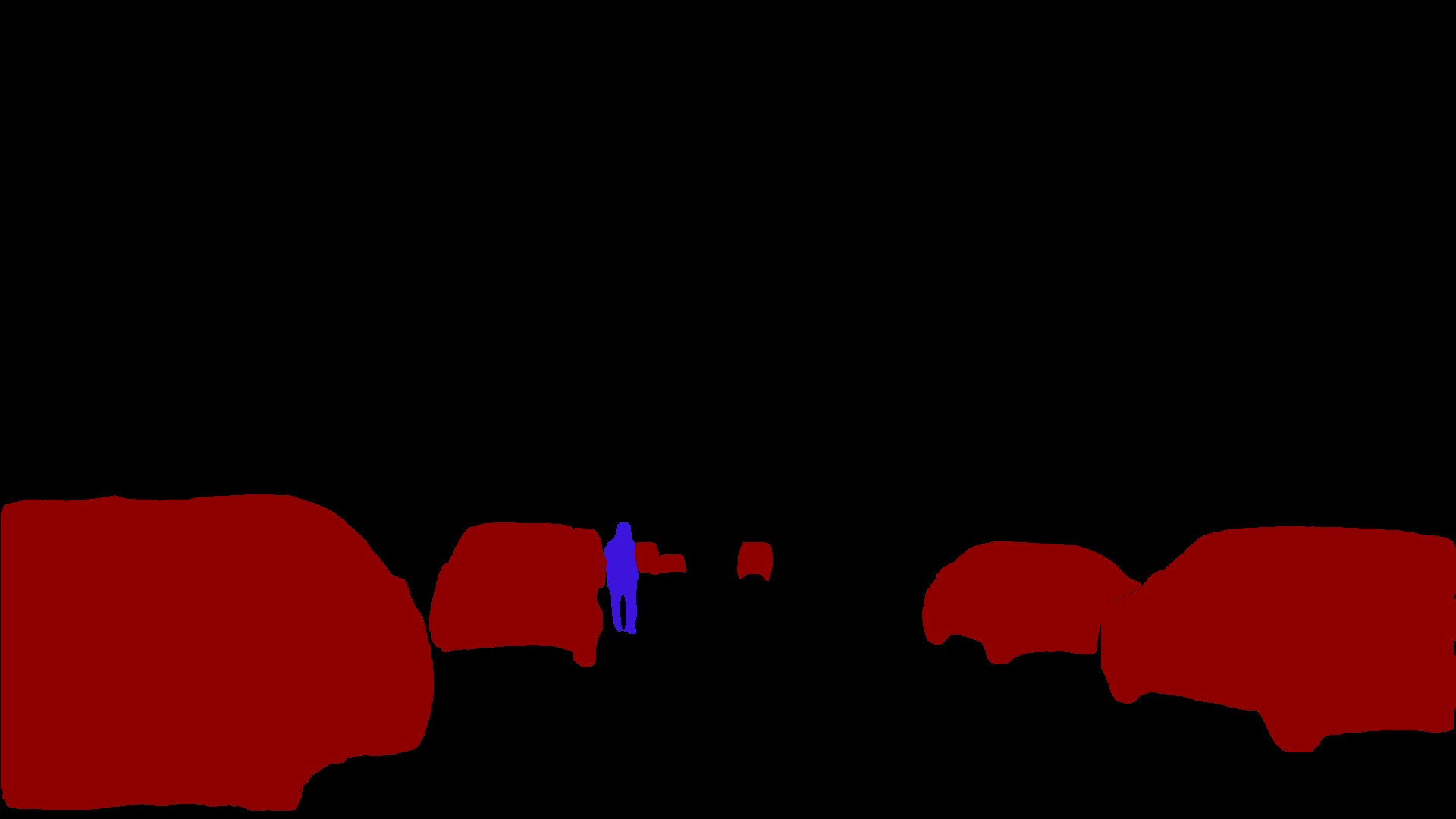} & 
\includegraphics[width=.32\columnwidth]{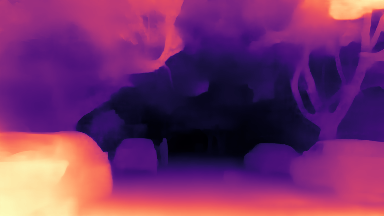}\\

\includegraphics[width=.32\columnwidth]{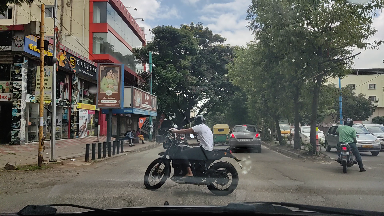} & 
\includegraphics[width=.32\columnwidth]{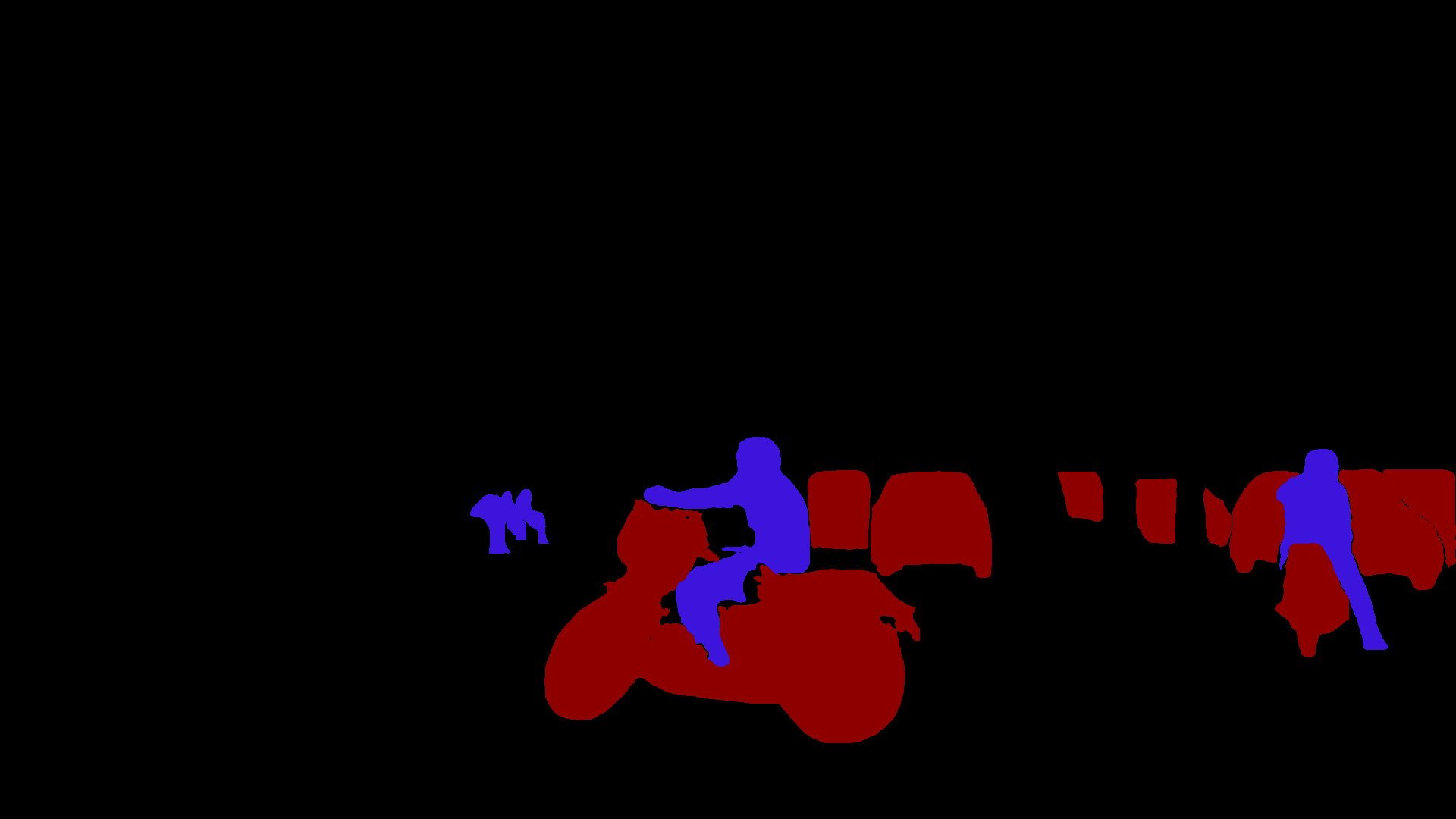} & 
\includegraphics[width=.32\columnwidth]{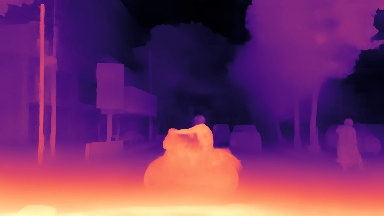}\\

\includegraphics[width=.32\columnwidth]{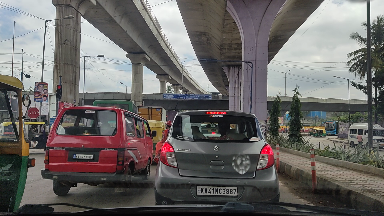} & 
\includegraphics[width=.32\columnwidth]{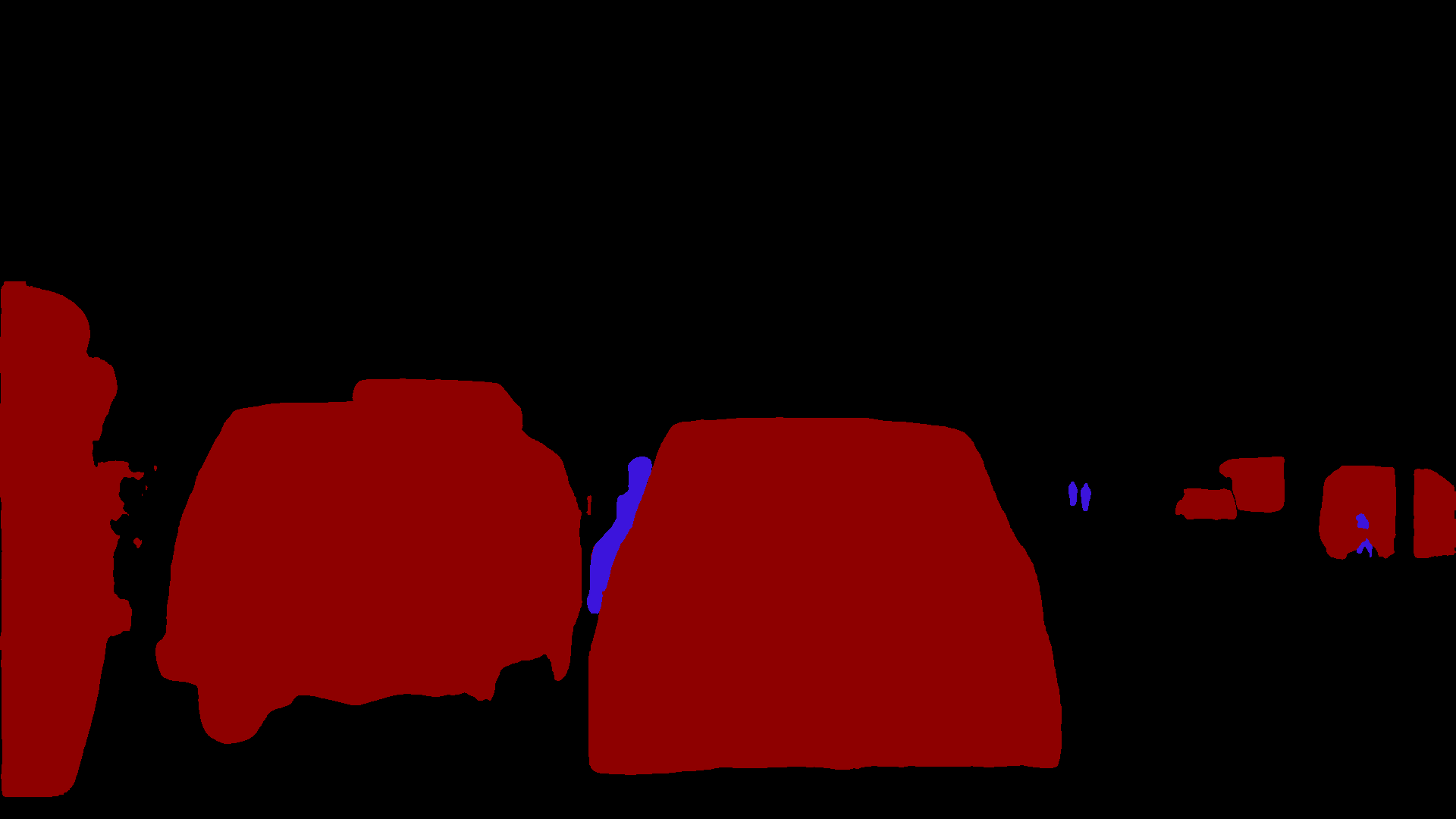} & 
\includegraphics[width=.32\columnwidth]{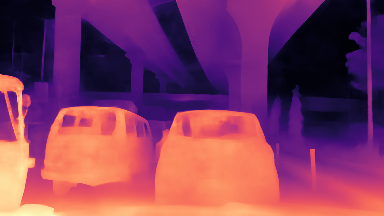}\\

\includegraphics[width=.32\columnwidth]{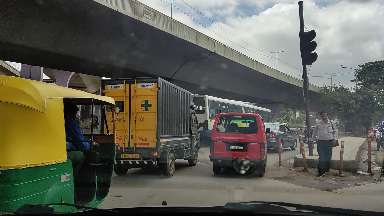} & 
\includegraphics[width=.32\columnwidth]{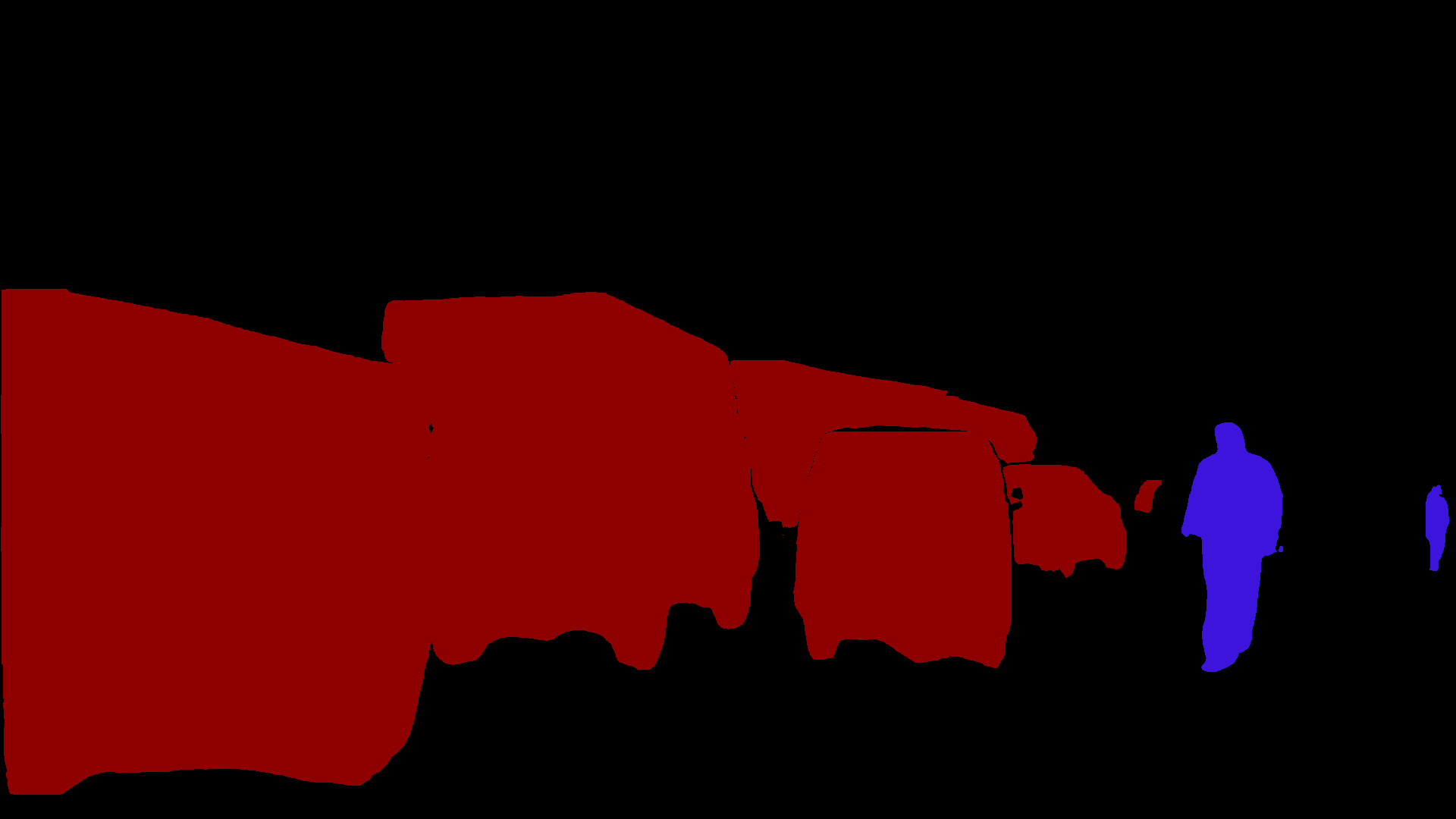} & 
\includegraphics[width=.32\columnwidth]{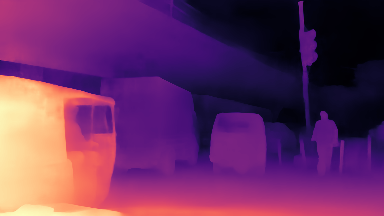}\\

\includegraphics[width=.32\columnwidth]{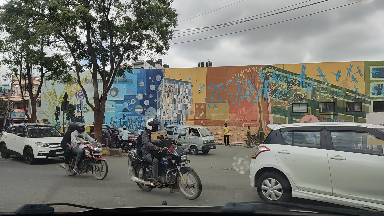} & 
\includegraphics[width=.32\columnwidth]{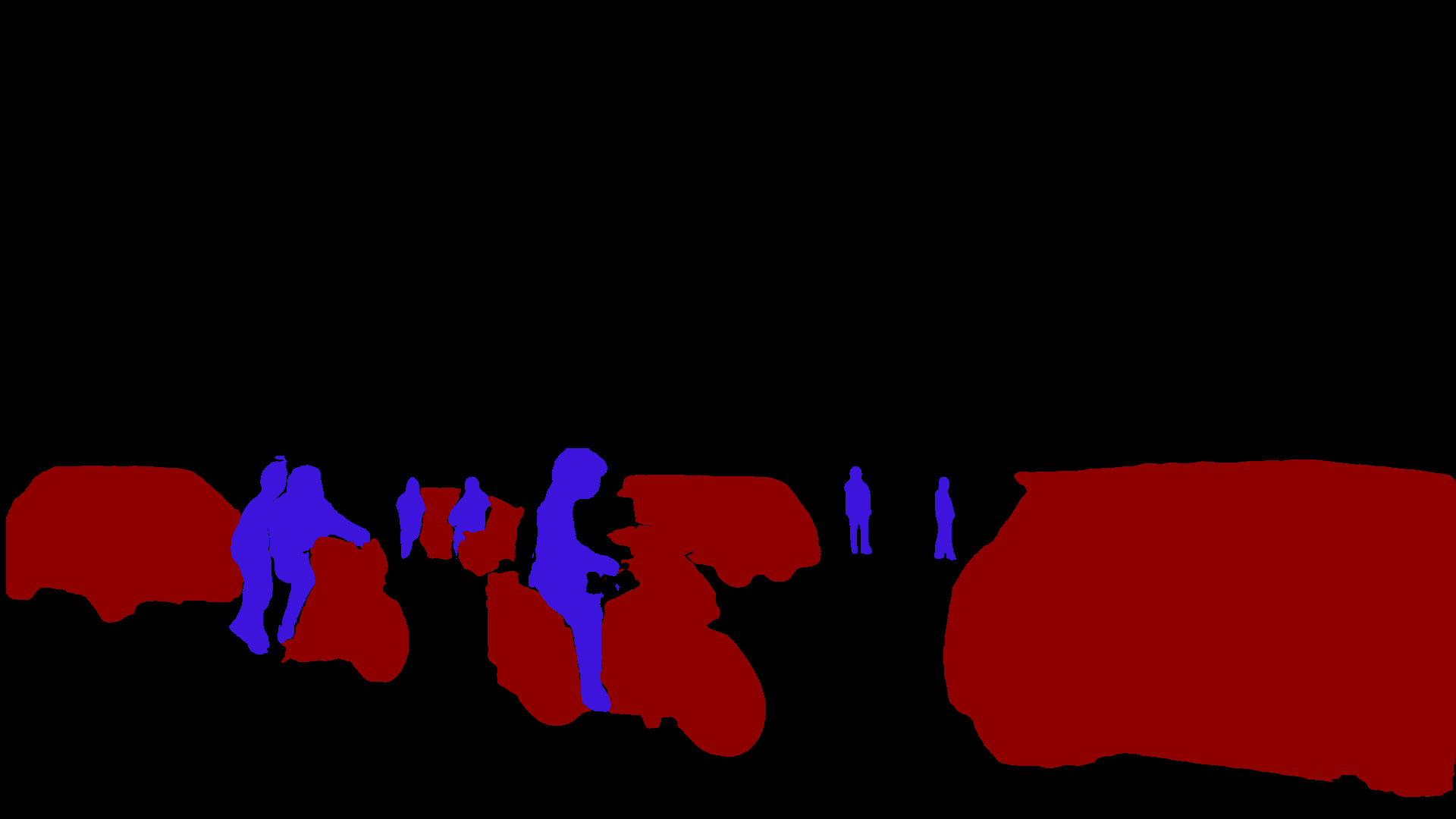} & 
\includegraphics[width=.32\columnwidth]{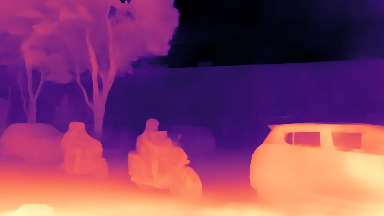}\\

  \end{tabular}
  \endgroup

  \hspace{0.5cm} \\

  \caption{We use Semantic Segmentation auto-labelling to generate semantic labels for the Bengaluru Driving Dataset. We have the RGB frames on the left, our segmentation maps in the middle and depth labels on the right. We would like to highlight the accuracy in the automatically generated segmentation maps}
  \label{fig:bdd_smenatic_labels} 
\end{figure}

\begin{figure*}[t]
\begin{center}
   \includegraphics[width=0.3\linewidth]{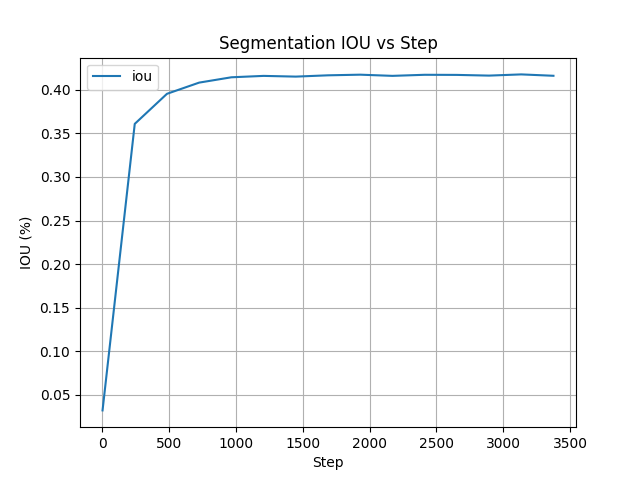}
   \includegraphics[width=0.3\linewidth]{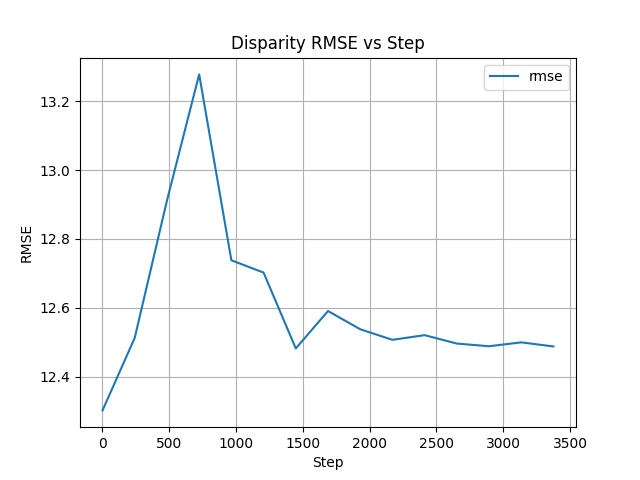}
   \includegraphics[width=0.3\linewidth]{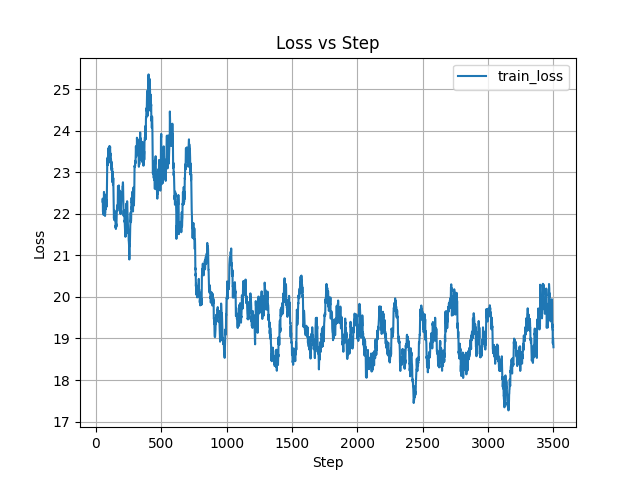} \\

   % \hspace{0.5cm} \\
   
   % \includegraphics[width=0.3\linewidth]{images/a1.png}
   % \includegraphics[width=0.3\linewidth]{images/a2.png}
   % \includegraphics[width=0.3\linewidth]{images/a3.png}
\end{center}
    % \hspace{0.5cm} \\
   \caption{While training $SOccDPT_{V3}$ we start with the pre-trained depth backbone. As a result, the initial disparity metrics (RMSE, a1, a2, a3) are good while the initial IoU score is under 5\%. Within the first few epochs the IoU score starts growing steadily, and we observe a small spike in the disparity metrics as the depth head is adjusting to the changes made to accommodate the segmentation head}
\label{fig:training_graph}
\end{figure*}

% The ability of vision based networks to learn and accurately predict based on image input is limited by their receptive field and their overall learning capacity (number of parameters). Most of these networks are trained on a predefined low input resolution. While it is possible to feed in larger input resolutions at test time, the system's performance will suffer due to the loss of features that would have otherwise been visible at higher resolutions. We use existing disparity estimation and image segmentation models to generate labels for existing datasets which don't already have the desired labels. The broad concept is to trade away compute time for higher accuracy as this allows us to generate pseudo-labels. The Indian Driving Dataset~\cite{IDD_DBLP:journals/corr/abs-1811-10200} has 2D semantic labels and we augment this dataset with depth labels. The Bengaluru Driving Dataset~\cite{OCTraN_analgund2023octran} has depth labels and we augment this dataset with 2D semantic labels.

The ability of vision based networks to learn and accurately predict based on image input is limited by their receptive field and their overall learning capacity (number of parameters). We generate pseudo-labels using existing disparity estimation and image segmentation models by feeding in segments of the image which allows the model to focus on smaller regions. This approach trades compute time for higher accuracy. We augment the Indian Driving Dataset~\cite{IDD_DBLP:journals/corr/abs-1811-10200} with depth labels and the Bengaluru Driving Dataset~\cite{OCTraN_analgund2023octran} with 2D semantic labels, enhancing their utility for training.

\textbf{Depth Boosting. } Monocular depth estimation systems use a lot of the depth cues used by humans including occlusion boundaries, parallel lines, edges, vanishing points and the shape and size of objects. Altering the resolution of the image affects the clarity of these depth cues. While increasing the resolution can produce sharper results, feeding in smaller patches of the image fails. This happens when the window size shrinks to the point where there are no depth cues, which generates an inconsistent overall structure and may introduce low frequency artefacts. Taking inspiration from the depth boosting techniques~\cite{OCTraN_analgund2023octran, Miangoleh2021Boosting, bmd_msc_boosting}, we select a content adaptive resolution $R_x$, beyond which the low frequency artefacts begin to hurt the overall structure of the image. By merging the disparity maps from the various resolutions, we are able to generate a high resolution disparity map with global consistency. As suggested by \cite{Miangoleh2021Boosting}, we select $R_{20}$ as the high-resolution upper bound $R_{x}$, as their work shows that using resolutions higher than $R_{30}$ results in a decrease in performance due to the aforementioned artefacts. We use this method to generate disparity labels for the Indian Driving Dataset as shown in ~\cref{fig:idd_depth_labels}. The depth images on the left are colored by inverse depth (or disparity), such that pixels representing objects closer to the camera are brighter and those representing objects further away are darker.

\textbf{Semantic Segmentation auto-labelling. } To produce high resolution 2D semantic labels, we take inspiration from PointRend~\cite{cheng2021pointly_pointrend}. We take an image as input and produce a coarse intermediate segmentation map using an existing segmentation approach MaskRCNN~\cite{mask_rcnn}. This coarse map is gradually up-sampled using bi-linear interpolation and only the regions of the resized map with high uncertainty are refined. The uncertain regions typically include the boundaries of objects. The uncertain region is refined by a lightweight multi-layered perceptron. Its input is a feature vector which is extracted through interpolation from the feature maps, which intern has been computed by the base model. As shown in ~\cref{fig:bdd_smenatic_labels} we have auto-labelled vehicles in red and humans in blue.

\section{EXPERIMENTS}

% \begin{figure}[t]
%     \centering
%     \includegraphics[width=.9\columnwidth]{images/iou.png}

%     \includegraphics[width=.9\columnwidth]{images/rmse.png}

%    \caption{While training $SOccDPT_{V3}$ we start with the pre-trained depth backbone. As a result, the initial RMSE is good while the initial IoU score is under 5\%. Within the first few epochs the IoU score starts growing steadily and we observe a small spike in the RMSE as the depth head is adjusting to the changes made to accommodate the segmentation head }
% \label{fig:training_graph}
% \end{figure}

\subsection{Experimental Setup}

\begin{table*}[]
    \vspace{0.5cm}    
    \centering
    \begin{tabular}{l|c|c|c|c|c|c|c}
        \toprule
        Model & Dataset & $RMSE^\downarrow$ & $a1^\uparrow$ & $a2^\uparrow$ & $a3^\uparrow$ & $FPS^\uparrow$ (Hz) & $Parameters^\downarrow$ \\
        \cmidrule(lr){1-8}
        $MiDaS_{v3.1}$ $Swin2_{T-256}$\cite{liu2021swinv2} &  BDD & 23.325 & 0.5944 & 0.7816 & 0.8585 & 82.2656 &  14.84M \\
        $MiDaS_{v3.0}$ $DPT_{H-384}$\cite{Ranftl2020_DPT} &  BDD & 21.861   & 0.5527 & 0.7712 & 0.8558 & 13.4394 & 123.1M\\
        $MiDaS_{v3.0}$ $DPT_{L-384}$\cite{Ranftl2020_DPT} &  BDD & 13.36    & 0.6888 & 0.8514 & 0.9192 &  6.3142 & 42.3M\\
        % $monodepth2$\cite{monodepth2} &  BDD & 37.6594    & 0.3602 & 0.5889 & 0.7304 &  \textbf{88.0492} & \textbf{14.84M}\\
        $manydepth$\cite{manydepth} &  BDD & 31.3599    & 0.4508 & 0.6668 & 0.7850 &  20.3578 & 35.34M\\
        $Zero Depth$\cite{tri-zerodepth} &  BDD & 36.3419    & 0.3882 & 0.6552 & 0.7875 &  2.363 & 232.59M\\
        $PackNet$\cite{tri-packnet} &  BDD & 50.6722    & 0.2257 & 0.4936 & 0.6870 &  8.7504 & 128.29M\\
        $SOccDPT_{V1}$ &  BDD & 13.3782 & 0.6854 & 0.8442 & 0.9172 & 39.1141 & 84.3M \\
        $SOccDPT_{V2}$ &  BDD & 26.2383 & 0.4879 & 0.7181 & 0.8309 & 69.6503 & 42.3M \\
        $SOccDPT_{V3}$ &  BDD & \textbf{12.4075} & \textbf{0.6935} & \textbf{0.8588} & \textbf{0.9265} & 69.4733 & 42.3M \\
        \bottomrule
    \end{tabular}

    \hspace{0.5cm} \\
    
    \caption{We compare SOccDPT's disparity metrics on the Bengaluru Driving Dataset, FPS and number of parameters with existing approaches. $SOccDPT_{V3}$ outperforms the models in terms of accuracy while maintaining a high FPS and small model size}
    \label{tab:other_methods}
\end{table*}

\begin{table*}[t]
    \centering
    \begin{tabular}{c l c c c c c c c c c}
        \toprule
        \multirow{2}{*}{\bfseries Method} & 
        \multirow{2}{*}{\bfseries Dataset} &
        \multicolumn{4}{c}{\bfseries Hyperparameters} &
        \multirow{2}{*}{\bfseries $RMSE^\downarrow$} &
        \multirow{1}{*}{\bfseries $a1^\uparrow$} &
        \multirow{1}{*}{\bfseries $a2^\uparrow$} &
        \multirow{1}{*}{\bfseries $a3^\uparrow$} &
        \multirow{2}{*}{\bfseries $IoU^\uparrow$ (\%)}
        %------ Multicolumn headings
        \\ \cmidrule(lr){3-6}
            && BS&PP&EP&LR & & $\delta<1.25^{1}$ & $\delta<1.25^{2}$ & $\delta<1.25^{3}$ &
        \\ \cmidrule(lr){1-11}
        %------
        % https://wandb.ai/pw22-sbn-01/depth_dpt_swin2_tiny_256/runs/v5jio69v
        % https://wandb.ai/pw22-sbn-01/seg_dpt_hybrid_384/runs/2iiafzb6
        \multirow{2}{*}{$SOccDPT_{V1}$} & IDD & 12* & 0.5 & 0.5 & 0.00001 &  11.2353 & 0.7717 & 0.8991 & 0.9211 & 42.48 \\
                                        & BDD & 12* & 0.5 & 0.5 & 0.00001 &  13.3782 & 0.6854 & 0.8442 & 0.9172 & 41.73 \\

        \multirow{2}{*}{$SOccDPT_{V2}$} & IDD & 6 & 0.5 & 0.95 & 0.00001 & 27.6473 & 0.5302 & 0.7084 & 0.8134 & 26.29 \\
                                        & BDD & 6 & 0.5 & 0.95 & 0.00001 & 26.2383 & 0.4879 & 0.7181 & 0.8309 & 34.75 \\
        \multirow{2}{*}{$SOccDPT_{V3}$} & IDD & 6 & 0.5 & 0.95 & 0.0001 & \textbf{9.1473} & \textbf{0.7807} & \textbf{0.9009} & \textbf{0.9416} & 43.50 \\
                                        & BDD & 6 & 0.5 & 0.95 & 0.0001 & 12.4075 & 0.6935 & 0.8588 & 0.9265 & \textbf{46.02}\\
        % SOccDPT_{V2} & IDD & 1 & 64 & 0 & 8 &  0.0001 & 32 & x & x & x\\
        \bottomrule
    \end{tabular}

    \hspace{0.5cm} \\
    
    \caption{\textbf{Ablation Study} SOccDPT's hyper-parameters and the metrics achieved. RMSE, a1, a2, a3 are disparity metrics and IoU is the segmentation metric. The hyper-parameters are batch size (BS), patch-wise percentage (PP), Encoder Percentage (EP) and learning rate (LR). Models with a * have had their two heads and backbones trained separately}
  \label{tab:modelComparision}
\end{table*}

We train SOccDPT on a laptop with an Intel i7-12700H (20 threads) and NVIDIA GeForce RTX 3070 Laptop GPU with 8 GB VRAM and utilize PyTorch 2.1.0~\cite{PyTorch_NEURIPS2019_9015}. With the goal of focusing performance in unstructured traffic, our network has been trained on the Indian Driving Dataset~\cite{IDD_DBLP:journals/corr/abs-1811-10200} and the Bengaluru Driving Dataset~\cite{OCTraN_analgund2023octran}. In ~\cref{tab:modelComparision}, we present the set of hyper-parameters which produce optimal results. We evaluate on the metrics Intersection over Union (IoU), Root Mean Squared Error (RMSE), threshold errors (\(a1\), \(a2\), \(a3\)). Here, \(a1\) is the fraction of predictions where the threshold \({\text{gt}} \slash {\text{pred}}\) or \({\text{pred}} \slash {\text{gt}}\) is less than \(1.25\); \(a2\) is the fraction of predictions where the threshold is less than \(1.25^2\) and \(a3\) is the fraction of predictions where the threshold is less than \(1.25^3\).

\subsection{Datasets}

\textbf{Indian Driving Dataset ~\cite{IDD_DBLP:journals/corr/abs-1811-10200} }. The IDD has a total of about 7974 frames with 6993 and 981 frames for training and testing respectively.

\textbf{Bengaluru Driving Dataset ~\cite{OCTraN_analgund2023octran}}. The BDD has a total of about 3629 frames. We split it to have 10\% for testing and the remainder for training.

\textbf{Bengaluru Semantic Occupancy Dataset (Ours)}. We extend BDD with 3D semantic occupancy labels by picking a voxel size of 50 cm and applied a voting filter to drop the voxels with fewer than 10 points. Since the disparity maps are in arbitrary scale, we manually estimated the scale of each frame.
% We settled on the occupancy grid size of $(256 \times 256 \times 32)$ voxels.

% \textbf{KITTI \cite{KITTI}}. TODO.

% \textbf{nuScenes \cite{fong2021_nue_panoptic}}. TODO.

\subsection{Ablation Study}
In ~\cref{tab:modelComparision}, we present the set of hyper-parameters for SOccDPT's $V1$, $V2$ and $V3$. We observe that $V1$ produces good disparity and segmentation metrics while also being the largest network in terms of number of parameters and the slowest to run as shown in ~\cref{tab:other_methods}. $SOccDPT_{V1}$ has two independent backbones which explains the larger number of parameters and increased inference time. While $SOccDPT_{V2}$ shows an improvement in speed and reduction in number of parameters, it takes a performance hit in terms of disparity and segmentation accuracy, as this network is being trained from scratch. $SOccDPT_{V2}$ introduces the common backbone which reduces the compute requirements, but since this entire network is being trained from scratch, it has no priors regarding either semantics or disparity estimation. We introduce this prior into $SOccDPT_{V3}$ by changing the architecture of $V2$ to allow us to load in pre-trained weights from the disparity backbone. As seen in ~\cref{fig:training_graph}, $SOccDPT_{V3}$ starts off with good RMSE scores for disparity estimation and poor IoU for segmentation, which is as expected. Through the course of training, the IoU steadily climbs. Initially, we see a spike in RMSE which comes back down over several epochs. $SOccDPT_{V3}$ has similar timing and memory characteristics when compared to $SOccDPT_{V2}$ as it is only a minor modification that allows us to load in the disparity backbone. But this small change allows $SOccDPT_{V3}$ to vastly outperform $SOccDPT_{V2}$ without requiring additional training data.

\subsection{Comparison with Existing Methods}

$SOccDPT_{V3}$'s performance exceeds existing disparity estimation approaches on unstructured traffic scenarios presented from the Bengaluru Driving Dataset. As shown in ~\cref{tab:other_methods}, $SOccDPT_{V3}$ shows the best accuracy in disparity estimation while also maintaining a high FPS and keeping compute requirements low. As shown in \cref{fig:qualitative_results}, our model provides very detailed disparity maps compared to existing approaches while performing in real time and keeping memory requirements low.

\section{CONCLUSIONS}

Existing disparity and segmentation approaches have come far, but do not specifically address the challenge in the autonomous vehicle context in unstructured traffic scenarios. We use depth boosting and semantic auto-labelling to build a self-supervised training pipeline, which can take videos as input and train a 3D semantic occupancy network. $SOccDPT$ uses a multi-headed Dense Transformer based architecture to take advantage of this self-supervised pipeline, to learn 3D semantic occupancy in the context of autonomous navigation in unstructured traffic. Our PatchWise training system allowed us to explore training with larger batch sizes which would not have been possible with memory constrained hardware. These models show potential in their ability to learn 3D semantic occupancy from monocular vision and operate at real time.

\section*{ACKNOWLEDGMENT}

We thank Dhruval Pobbathi Badrinath for the fruitful discussions through the course of the project.

%%%%%%%%%%%%%%%%%%%%%%%%%%%%%%%%%%%%%%%%%%%%%%%%%%%%%%%%%%%%%%%%%%%%%%%%%%%%%%%%

{\small
\bibliographystyle{IEEEtran}
\bibliography{root}

\begin{thebibliography}{10}
\providecommand{\url}[1]{#1}
\csname url@samestyle\endcsname
\providecommand{\newblock}{\relax}
\providecommand{\bibinfo}[2]{#2}
\providecommand{\BIBentrySTDinterwordspacing}{\spaceskip=0pt\relax}
\providecommand{\BIBentryALTinterwordstretchfactor}{4}
\providecommand{\BIBentryALTinterwordspacing}{\spaceskip=\fontdimen2\font plus
\BIBentryALTinterwordstretchfactor\fontdimen3\font minus
  \fontdimen4\font\relax}
\providecommand{\BIBforeignlanguage}[2]{{%
\expandafter\ifx\csname l@#1\endcsname\relax
\typeout{** WARNING: IEEEtran.bst: No hyphenation pattern has been}%
\typeout{** loaded for the language `#1'. Using the pattern for}%
\typeout{** the default language instead.}%
\else
\language=\csname l@#1\endcsname
\fi
#2}}
\providecommand{\BIBdecl}{\relax}
\BIBdecl

\bibitem{OCTraN_analgund2023octran}
\BIBentryALTinterwordspacing
A.~N. Ganesh, D.~Pobbathi~Badrinath, H.~M. Kumar, P.~S, and S.~Narayan,
  ``Octran: 3d occupancy convolutional transformer network in unstructured
  traffic scenarios,'' Spotlight Presentation at the Transformers for Vision
  Workshop, CVPR, 2023, transformers for Vision Workshop, CVPR 2023. [Online].
  Available: \url{https://sites.google.com/view/t4v-cvpr23/papers}
\BIBentrySTDinterwordspacing

\bibitem{bao2022beit}
\BIBentryALTinterwordspacing
H.~Bao, L.~Dong, S.~Piao, and F.~Wei, ``{BE}it: {BERT} pre-training of image
  transformers,'' in \emph{International Conference on Learning
  Representations}, 2022. [Online]. Available:
  \url{https://openreview.net/forum?id=p-BhZSz59o4}
\BIBentrySTDinterwordspacing

\bibitem{Graham_2021_ICCV_levit}
B.~Graham, A.~El-Nouby, H.~Touvron, P.~Stock, A.~Joulin, H.~Jegou, and
  M.~Douze, ``Levit: A vision transformer in convnet's clothing for faster
  inference,'' in \emph{Proceedings of the IEEE/CVF International Conference on
  Computer Vision (ICCV)}, October 2021, pp. 12\,259--12\,269.

\bibitem{jaegle2021perceiver}
A.~Jaegle, S.~Borgeaud, J.-B. Alayrac, C.~Doersch, C.~Ionescu, D.~Ding,
  S.~Koppula, A.~Brock, E.~Shelhamer, O.~Hénaff, M.~M. Botvinick,
  A.~Zisserman, O.~Vinyals, and J.~Carreira, ``Perceiver io: A general
  architecture for structured inputs \& outputs,'' 2021.

\bibitem{li2022next_vit}
J.~Li, X.~Xia, W.~Li, H.~Li, X.~Wang, X.~Xiao, R.~Wang, M.~Zheng, and X.~Pan,
  ``Next-vit: Next generation vision transformer for efficient deployment in
  realistic industrial scenarios,'' \emph{arXiv preprint arXiv:2207.05501},
  2022.

\bibitem{liu2021swinv2}
Z.~Liu, H.~Hu, Y.~Lin, Z.~Yao, Z.~Xie, Y.~Wei, J.~Ning, Y.~Cao, Z.~Zhang,
  L.~Dong, F.~Wei, and B.~Guo, ``Swin transformer v2: Scaling up capacity and
  resolution,'' in \emph{International Conference on Computer Vision and
  Pattern Recognition (CVPR)}, 2022.

\bibitem{liu2021Swin}
Z.~Liu, Y.~Lin, Y.~Cao, H.~Hu, Y.~Wei, Z.~Zhang, S.~Lin, and B.~Guo, ``Swin
  transformer: Hierarchical vision transformer using shifted windows,'' in
  \emph{Proceedings of the IEEE/CVF International Conference on Computer Vision
  (ICCV)}, 2021.

\bibitem{ViT}
A.~Dosovitskiy, L.~Beyer, A.~Kolesnikov, D.~Weissenborn, X.~Zhai,
  T.~Unterthiner, M.~Dehghani, M.~Minderer, G.~Heigold, S.~Gelly \emph{et~al.},
  ``An image is worth 16x16 words: Transformers for image recognition at
  scale,'' \emph{arXiv preprint arXiv:2010.11929}, 2020.

\bibitem{fong2021_nue_panoptic}
W.~K. Fong, R.~Mohan, J.~V. Hurtado, L.~Zhou, H.~Caesar, O.~Beijbom, and
  A.~Valada, ``Panoptic nuscenes: A large-scale benchmark for lidar panoptic
  segmentation and tracking,'' \emph{arXiv preprint arXiv:2109.03805}, 2021.

\bibitem{KITTI}
A.~Geiger, P.~Lenz, C.~Stiller, and R.~Urtasun, ``Vision meets robotics: the
  kitti dataset,'' \emph{The International Journal of Robotics Research},
  vol.~32, pp. 1231--1237, 09 2013.

\bibitem{KITTI_360_DBLP:journals/corr/abs-2109-13410}
\BIBentryALTinterwordspacing
Y.~Liao, J.~Xie, and A.~Geiger, ``{KITTI-360:} {A} novel dataset and benchmarks
  for urban scene understanding in 2d and 3d,'' \emph{CoRR}, vol.
  abs/2109.13410, 2021. [Online]. Available:
  \url{https://arxiv.org/abs/2109.13410}
\BIBentrySTDinterwordspacing

\bibitem{cheng2021pointly_pointrend}
B.~Cheng, O.~Parkhi, and A.~Kirillov, ``Pointly-supervised instance
  segmentation,'' 2021.

\bibitem{bmd_msc_boosting}
S.~M.~H. Miangoleh, ``Boosting monocular depth estimation to high resolution,''
  Master's thesis, Simon Fraser University, 2022.

\bibitem{Miangoleh2021Boosting}
S.~M.~H. Miangoleh, S.~Dille, L.~Mai, S.~Paris, and Y.~Aksoy, ``Boosting
  monocular depth estimation models to high-resolution via content-adaptive
  multi-resolution merging,'' in \emph{2021 IEEE/CVF Conference on Computer
  Vision and Pattern Recognition (CVPR)}, 2021, pp. 9680--9689.

\bibitem{IDD_DBLP:journals/corr/abs-1811-10200}
\BIBentryALTinterwordspacing
G.~Varma, A.~Subramanian, A.~M. Namboodiri, M.~Chandraker, and C.~V. Jawahar,
  ``{IDD:} {A} dataset for exploring problems of autonomous navigation in
  unconstrained environments,'' \emph{CoRR}, vol. abs/1811.10200, 2018.
  [Online]. Available: \url{http://arxiv.org/abs/1811.10200}
\BIBentrySTDinterwordspacing

\bibitem{Ranftl2022_midas_ssi_loss}
R.~Ranftl, K.~Lasinger, D.~Hafner, K.~Schindler, and V.~Koltun, ``Towards
  robust monocular depth estimation: Mixing datasets for zero-shot
  cross-dataset transfer,'' \emph{IEEE Transactions on Pattern Analysis and
  Machine Intelligence}, vol.~44, no.~3, 2022.

\bibitem{Liu2018ERN}
S.~Liu, W.~Ding, C.~Liu, Y.~Liu, and H.~Li, ``Ern: Edge loss reinforced
  semantic segmentation network for remote sensing images,'' \emph{Remote
  Sensing}, vol.~10, no.~9, p. 1339, 2018.

\bibitem{joint_semantic_mtl}
M.~Zhen, J.~Wang, L.~Zhou, S.~Li, T.~Shen, S.~Jiaxiang, T.~Fang, and L.~Quan,
  ``Joint semantic segmentation and boundary detection using iterative pyramid
  contexts,'' 06 2020, pp. 13\,663--13\,672.

\bibitem{9706962_multi_task_ssl}
Y.~Wang, Y.-H. Tsai, W.-C. Hung, W.~Ding, S.~Liu, and M.-H. Yang,
  ``Semi-supervised multi-task learning for semantics and depth,'' in
  \emph{2022 IEEE/CVF Winter Conference on Applications of Computer Vision
  (WACV)}, 2022, pp. 2663--2672.

\bibitem{mtl_survey}
S.~Vandenhende, S.~Georgoulis, W.~Gansbeke, M.~Proesmans, D.~Dai, and L.~Gool,
  ``Multi-task learning for dense prediction tasks: A survey,'' \emph{IEEE
  Transactions on Pattern Analysis and Machine Intelligence}, vol.~PP, pp.
  1--1, 01 2021.

\bibitem{Cheng2016SemiSupervisedMD}
Y.~Cheng, X.~Zhao, R.~Cai, Z.~Li, K.~Huang, and Y.~Rui, ``Semi-supervised
  multimodal deep learning for rgb-d object recognition,'' in
  \emph{International Joint Conference on Artificial Intelligence}, 2016.

\bibitem{6963417_ssl_mtl_scene}
X.~Lu, X.~Li, and L.~Mou, ``Semi-supervised multitask learning for scene
  recognition,'' \emph{IEEE Transactions on Cybernetics}, vol.~45, no.~9, pp.
  1967--1976, 2015.

\bibitem{monodepth}
C.~Godard, O.~Aodha, and G.~Brostow, ``Unsupervised monocular depth estimation
  with left-right consistency,'' 07 2017.

\bibitem{monodepth2}
C.~Godard, O.~Aodha, M.~Firman, and G.~Brostow, ``Digging into self-supervised
  monocular depth estimation,'' 11 2019.

\bibitem{megadepth}
Z.~Li and N.~Snavely, ``Megadepth: Learning single-view depth prediction from
  internet photos,'' in \emph{Proceedings of the IEEE Conference on Computer
  Vision and Pattern Recognition (CVPR)}, June 2018.

\bibitem{gcndepth}
A.~Masoumian, H.~A. Rashwan, S.~Abdulwahab, J.~Cristiano, M.~S. Asif, and
  D.~Puig, ``Gcndepth: Self-supervised monocular depth estimation based on
  graph convolutional network,'' \emph{Neurocomputing}, 2022.

\bibitem{manydepth}
\BIBentryALTinterwordspacing
J.~Watson, O.~M. Aodha, V.~Prisacariu, G.~Brostow, and M.~Firman, ``The
  temporal opportunist: Self-supervised multi-frame monocular depth,'' in
  \emph{2021 IEEE/CVF Conference on Computer Vision and Pattern Recognition
  (CVPR)}.\hskip 1em plus 0.5em minus 0.4em\relax Los Alamitos, CA, USA: IEEE
  Computer Society, June 2021, pp. 1164--1174. [Online]. Available:
  \url{https://doi.ieeecomputersociety.org/10.1109/CVPR46437.2021.00122}
\BIBentrySTDinterwordspacing

\bibitem{hung2018adversarial_pseudo_label}
\BIBentryALTinterwordspacing
W.-C. Hung, Y.-H. Tsai, Y.-T. Liou, Y.-Y. Lin, and M.-H. Yang, ``Adversarial
  learning for semi-supervised semantic segmentation,'' 2018. [Online].
  Available: \url{https://openreview.net/forum?id=SJQO7UJCW}
\BIBentrySTDinterwordspacing

\bibitem{NIPS2004_96f2b50b_entropy_min}
\BIBentryALTinterwordspacing
Y.~Grandvalet and Y.~Bengio, ``Semi-supervised learning by entropy
  minimization,'' in \emph{Advances in Neural Information Processing Systems},
  L.~Saul, Y.~Weiss, and L.~Bottou, Eds., vol.~17.\hskip 1em plus 0.5em minus
  0.4em\relax MIT Press, 2004. [Online]. Available:
  \url{https://proceedings.neurips.cc/paper_files/paper/2004/file/96f2b50b5d3613adf9c27049b2a888c7-Paper.pdf}
\BIBentrySTDinterwordspacing

\bibitem{cheng2020hierarchical}
X.~Cheng, Y.~Zhong, M.~Harandi, Y.~Dai, X.~Chang, H.~Li, T.~Drummond, and
  Z.~Ge, ``Hierarchical neural architecture search for deep stereo matching,''
  \emph{Advances in Neural Information Processing Systems}, vol.~33, 2020.

\bibitem{UASNet}
Y.~Mao, Z.~Liu, W.~Li, Y.~Dai, Q.~Wang, Y.-T. Kim, and H.-S. Lee, ``Uasnet:
  Uncertainty adaptive sampling network for deep stereo matching,'' in
  \emph{2021 IEEE/CVF International Conference on Computer Vision (ICCV)},
  2021, pp. 6291--6299.

\bibitem{UPFNet}
Z.~Wu, X.~Wu, X.~Zhang, S.~Wang, and L.~Ju, ``Semantic stereo matching with
  pyramid cost volumes,'' in \emph{2019 IEEE/CVF International Conference on
  Computer Vision (ICCV)}, 2019, pp. 7483--7492.

\bibitem{xu2023iterative_IGEVStereo}
G.~Xu, X.~Wang, X.~Ding, and X.~Yang, ``Iterative geometry encoding volume for
  stereo matching,'' in \emph{Proceedings of the IEEE/CVF Conference on
  Computer Vision and Pattern Recognition}, 2023, pp. 21\,919--21\,928.

\bibitem{li2022bevformer}
Z.~Li, W.~Wang, H.~Li, E.~Xie, C.~Sima, T.~Lu, Y.~Qiao, and J.~Dai,
  ``Bevformer: Learning bird’s-eye-view representation from multi-camera
  images via spatiotemporal transformers,'' \emph{arXiv preprint
  arXiv:2203.17270}, 2022.

\bibitem{reiher2020sim2real}
L.~Reiher, B.~Lampe, and L.~Eckstein, ``A sim2real deep learning approach for
  the transformation of images from multiple vehicle-mounted cameras to a
  semantically segmented image in bird’s eye view,'' in \emph{2020 IEEE 23rd
  International Conference on Intelligent Transportation Systems (ITSC)}, 2020,
  pp. 1--7.

\bibitem{roddick2020predicting}
T.~Roddick and R.~Cipolla, ``Predicting semantic map representations from
  images using pyramid occupancy networks,'' 06 2020, pp. 11\,135--11\,144.

\bibitem{liu2022bevfusion}
Z.~Liu, H.~Tang, A.~Amini, X.~Yang, H.~Mao, D.~Rus, and S.~Han, ``Bevfusion:
  Multi-task multi-sensor fusion with unified bird's-eye view representation,''
  in \emph{IEEE International Conference on Robotics and Automation (ICRA)},
  2023.

\bibitem{zhu2020cylindrical}
\BIBentryALTinterwordspacing
X.~Zhu, H.~Zhou, T.~Wang, F.~Hong, Y.~Ma, W.~Li, H.~Li, and D.~Lin,
  ``Cylindrical and asymmetrical 3d convolution networks for lidar
  segmentation,'' in \emph{2021 IEEE/CVF Conference on Computer Vision and
  Pattern Recognition (CVPR)}.\hskip 1em plus 0.5em minus 0.4em\relax Los
  Alamitos, CA, USA: IEEE Computer Society, June 2021, pp. 9934--9943.
  [Online]. Available:
  \url{https://doi.ieeecomputersociety.org/10.1109/CVPR46437.2021.00981}
\BIBentrySTDinterwordspacing

\bibitem{zhou2017voxelnet}
Y.~Zhou and O.~Tuzel, ``Voxelnet: End-to-end learning for point cloud based 3d
  object detection,'' in \emph{2018 IEEE/CVF Conference on Computer Vision and
  Pattern Recognition}, 2018, pp. 4490--4499.

\bibitem{cheng2021af2s3net}
R.~Cheng, R.~Razani, E.~Taghavi, T.~Li, and B.~Liu, ``(af) 2 -s3net: Attentive
  feature fusion with adaptive feature selection for sparse semantic
  segmentation network,'' 06 2021, pp. 12\,542--12\,551.

\bibitem{liong2020amvnet}
V.~E. Liong, T.~N.~T. Nguyen, S.~Widjaja, D.~Sharma, and Z.~J. Chong, ``Amvnet:
  Assertion-based multi-view fusion network for lidar semantic segmentation,''
  2020.

\bibitem{tang2020searching}
\BIBentryALTinterwordspacing
H.~Tang, Z.~Liu, S.~Zhao, Y.~Lin, J.~Lin, H.~Wang, and S.~Han, ``Searching
  efficient 3d architectures with sparse point-voxel convolution,'' in
  \emph{Computer Vision - {ECCV} 2020 - 16th European Conference, Glasgow, UK,
  August 23-28, 2020, Proceedings, Part {XXVIII}}, ser. Lecture Notes in
  Computer Science, A.~Vedaldi, H.~Bischof, T.~Brox, and J.~Frahm, Eds., vol.
  12373.\hskip 1em plus 0.5em minus 0.4em\relax Springer, 2020, pp. 685--702.
  [Online]. Available: \url{https://doi.org/10.1007/978-3-030-58604-1\_41}
\BIBentrySTDinterwordspacing

\bibitem{ye2022lidarmultinet}
\BIBentryALTinterwordspacing
D.~Ye, Z.~Zhou, W.~Chen, Y.~Xie, Y.~Wang, P.~Wang, and H.~Foroosh,
  ``Lidarmultinet: Towards a unified multi-task network for lidar perception,''
  \emph{CoRR}, vol. abs/2209.09385, 2022. [Online]. Available:
  \url{https://doi.org/10.48550/arXiv.2209.09385}
\BIBentrySTDinterwordspacing

\bibitem{ye2021drinet}
M.~Ye, R.~Wan, S.~Xu, T.~Cao, and Q.~Chen, ``Drinet++: Efficient voxel-as-point
  point cloud segmentation,'' 2021.

\bibitem{cao2022monoscene}
A.-Q. Cao and R.~de~Charette, ``Monoscene: Monocular 3d semantic scene
  completion,'' in \emph{CVPR}, 2022.

\bibitem{chen20203d}
X.~Chen, K.-Y. Lin, C.~Qian, G.~Zeng, and H.~Li, ``3d sketch-aware semantic
  scene completion via semi-supervised structure prior,'' \emph{2020 IEEE/CVF
  Conference on Computer Vision and Pattern Recognition (CVPR)}, pp.
  4192--4201, 2020.

\bibitem{li2020anisotropic}
J.~Li, K.~Han, P.~Wang, Y.~Liu, and X.~Yuan, ``Anisotropic convolutional
  networks for 3d semantic scene completion,'' 06 2020, pp. 3348--3356.

\bibitem{roldão2020lmscnet}
L.~Roldão, R.~de~Charette, and A.~Verroust-Blondet, ``Lmscnet: Lightweight
  multiscale 3d semantic completion,'' in \emph{2020 International Conference
  on 3D Vision (3DV)}, 2020, pp. 111--119.

\bibitem{yan2020sparse}
X.~Yan, J.~Gao, J.~Li, R.~Zhang, Z.~Li, R.~Huang, and S.~Cui, ``Sparse single
  sweep lidar point cloud segmentation via learning contextual shape priors
  from scene completion,'' in \emph{AAAI Conference on Artificial
  Intelligence}, 2020.

\bibitem{huang2023triperspective}
Y.~Huang, W.~Zheng, Y.~Zhang, J.~Zhou, and J.~Lu, ``Tri-perspective view for
  vision-based 3d semantic occupancy prediction,'' \emph{arXiv preprint
  arXiv:2302.07817}, 2023.

\bibitem{Ranftl2020_DPT}
R.~Ranftl, K.~Lasinger, D.~Hafner, K.~Schindler, and V.~Koltun, ``Towards
  robust monocular depth estimation: Mixing datasets for zero-shot
  cross-dataset transfer,'' \emph{IEEE Transactions on Pattern Analysis and
  Machine Intelligence (TPAMI)}, 2020.

\bibitem{tri-packnet}
V.~Guizilini, R.~Ambrus, S.~Pillai, A.~Raventos, and A.~Gaidon, ``3d packing
  for self-supervised monocular depth estimation,'' in \emph{Proceedings of the
  International Conference on Computer Vision and Pattern Recognition (CVPR)},
  2020.

\bibitem{tri-zerodepth}
V.~Guizilini, I.~Vasiljevic, D.~Chen, R.~Ambrus, and A.~Gaidon, ``Towards
  zero-shot scale-aware monocular depth estimation,'' in \emph{Proceedings of
  the IEEE/CVF International Conference on Computer Vision (ICCV)}, October
  2023.

\bibitem{PyTorch_NEURIPS2019_9015}
A.~Paszke, S.~Gross, F.~Massa, A.~Lerer, J.~Bradbury, G.~Chanan, T.~Killeen,
  Z.~Lin, N.~Gimelshein, L.~Antiga, A.~Desmaison, A.~Kopf, E.~Yang, Z.~DeVito,
  M.~Raison, A.~Tejani, S.~Chilamkurthy, B.~Steiner, L.~Fang, J.~Bai, and
  S.~Chintala, ``Pytorch: An imperative style, high-performance deep learning
  library,'' in \emph{Advances in Neural Information Processing Systems
  32}.\hskip 1em plus 0.5em minus 0.4em\relax Curran Associates, Inc., 2019,
  pp. 8024--8035.

\bibitem{mask_rcnn}
K.~He, G.~Gkioxari, P.~Dollar, and R.~Girshick, ``Mask r-cnn,'' 10 2017, pp.
  2980--2988.

\end{thebibliography}
}

\end{document}